%% file: main.tex
\documentclass{article}

\usepackage{microtype}
\usepackage{graphicx}
\usepackage{booktabs, multirow} 
\usepackage[normalem]{ulem}

\usepackage{hyperref}

\usepackage{algorithm}
\usepackage{algpseudocode}

\input{math_commands.tex}

\usepackage[accepted]{icml2024}

\usepackage{amsmath}
\usepackage{amssymb}
\usepackage{mathtools}
\usepackage{amsthm}
\usepackage{hyperref}

\usepackage[capitalize,noabbrev]{cleveref}

\theoremstyle{plain}

\theoremstyle{definition}

\theoremstyle{remark}

\usepackage[textsize=tiny]{todonotes}

\usepackage{times}  
\usepackage{helvet}  
\usepackage{courier}  
\usepackage{graphicx} 
\usepackage{caption} 

\usepackage{subfigure}

\usepackage{amssymb}
\usepackage{caption,subcaption}
\usepackage{amsmath}
\usepackage{multirow}
\usepackage{xcolor}
\usepackage{bbold}
\usepackage{colortbl}
\usepackage{makecell}
\usepackage{newfloat}
\usepackage{listings}
\usepackage{hyperref}
\definecolor{customcolor}{HTML}{004496}
\hypersetup{
colorlinks=true,
linkcolor=red,
citecolor=customcolor
}
\usepackage{url}
\usepackage{wrapfig}

\definecolor{airforceblue}{rgb}{0.14, 0.31, 0.5}
\definecolor{brightmaroon}{rgb}{0.76, 0.13, 0.28}
\definecolor{mdgreen}{rgb}{0.05,0.6,0.05}
\definecolor{mdairforceblue}{rgb}{0.1, 0.3, 0.7}
\definecolor{mdblue}{rgb}{0,0,0.7}

\newcommand{\smallred}[1]{\small \textcolor{brightmaroon}{#1}}
\newcommand{\smallgreen}[1]{\small \textcolor{mdgreen}{#1}}

\definecolor{orange}{rgb}{1,0.5,0}
\definecolor{dkblue}{rgb}{0,0,0.5}
\definecolor{dkgray}{rgb}{0.3,0.3,0.3}
\definecolor{slate}{rgb}{0.25,0.25,0.4}
\definecolor{gray}{rgb}{0.5,0.5,0.5}
\definecolor{ltgray}{rgb}{0.7,0.7,0.7}
\definecolor{purple}{rgb}{0.7,0,1.0}
\definecolor{lavender}{rgb}{0.65,0.55,1.0}

\newcommand{\ie}{\textit{i}.\textit{e}.}
\newcommand{\eg}{\textit{e}.\textit{g}.} 

\newcommand{\xr}{X_{\mathrm{r}}}
\newcommand{\xc}{X_{\mathrm{c}}}
\newcommand{\A}{\texttt{\{A\}}}
\newcommand{\B}{\texttt{\{B\}}}
\newcommand{\C}{\texttt{\{C\}}}

\definecolor{lightgrey}{RGB}{244,244,244}
\definecolor{grey}{RGB}{128,128,128}
\definecolor{midgrey}{RGB}{225,225,225}
\definecolor{forestgreen}{RGB}{47, 159, 87}

\usepackage{colortbl}  
\usepackage{array}
\definecolor{Gray}{gray}{0.94}
\definecolor{liGray}{gray}{0.5}
\definecolor{LightCyan}{rgb}{0.88,1,1}

\icmltitlerunning{Interpreting and Improving Large Language Models in Arithmetic Calculation}

\begin{document}

\twocolumn[
\icmltitle{Interpreting and Improving Large Language Models in Arithmetic Calculation}

\icmlsetsymbol{intern}{*}
\icmlsetsymbol{corr}{$\dagger$}

\begin{icmlauthorlist}
\icmlauthor{Wei Zhang}{intern,USTC,Ali}
\icmlauthor{Chaoqun Wan}{Ali}
\icmlauthor{Yonggang Zhang}{corr,HK}
\icmlauthor{Yiu-ming Cheung}{HK}
\icmlauthor{Xinmei Tian}{USTC,HF}
\icmlauthor{Xu Shen}{corr,Ali}
\icmlauthor{Jieping Ye}{Ali}
\end{icmlauthorlist}

\icmlaffiliation{USTC}{University of Science and Technology of China}
\icmlaffiliation{Ali}{Alibaba Cloud}
\icmlaffiliation{HK}{Hong Kong Baptist University}
\icmlaffiliation{HF}{Institute of Artificial Intelligence, Hefei Comprehensive National Science Center}

\icmlcorrespondingauthor{Xu Shen$^{\dagger}$}{shenxu.sx@alibaba-inc.com}
\icmlcorrespondingauthor{Yonggang Zhang$^{\dagger}$}{csygzhang@comp.hkbu.edu.hk}

\icmlkeywords{Machine Learning, ICML}

\vskip 0.3in
]

\printAffiliationsAndNotice{{\textsuperscript{*}This work was done when the author was visiting Alibaba Cloud as a research intern.}}

\begin{abstract}
\input{tex/0_abs}
\end{abstract}

\section{Introduction}
\label{sec:intro}
\input{tex/1_intro}

\section{Related Works}
\label{sec:exp}
\input{tex/5_related_works}

\section{Preliminary}
\label{sec:related}
\input{tex/2_background}

\section{Method}
\label{sec:method}
\input{tex/3_method}

\section{Experiments}
\label{sec:exp}
\input{tex/4_exp}

\section{Conclusion}
\label{sec:conclusion}
\input{tex/7_conclusion}

\section*{Acknowledgements}
\noindent
This work was supported in part by NSFC No. 62222117.
YGZ and YMC were supported in part by NSFC/Research Grants Council (RGC) Joint Research Scheme under Grant: N\_HKBU214/21; in part by RGC Senior Research Fellow Scheme under Grant: SRFS2324-2S02.

\section*{Impact Statement}
This paper presents work whose goal is to advance the field of Machine Learning. There are many potential societal consequences of our work, none of which we feel must be specifically highlighted here.




\bibliography{main}
\bibliographystyle{icml2024}

\newpage
\appendix
\onecolumn

\input{tex/8_appendix}


\end{document}

%% file: math_commands.tex
\usepackage{amsmath,amsfonts,bm}

\def\eqref#1{equation~\ref{#1}}

\def\1{\bm{1}}

\DeclareMathAlphabet{\mathsfit}{\encodingdefault}{\sfdefault}{m}{sl}
\SetMathAlphabet{\mathsfit}{bold}{\encodingdefault}{\sfdefault}{bx}{n}



%% file: tex/0_abs.tex
Large language models (LLMs) have demonstrated remarkable potential across numerous applications and have shown an emergent ability to tackle complex reasoning tasks, such as mathematical computations. However, even for the simplest arithmetic calculations, the intrinsic mechanisms behind LLMs remain mysterious, making it challenging to ensure reliability. In this work, we delve into uncovering a specific mechanism by which LLMs execute calculations. Through comprehensive experiments, we find that LLMs frequently involve a small fraction ($<5\%$) of attention heads, which play a pivotal role in focusing on operands and operators during calculation processes. Subsequently, the information from these operands is processed through multi-layer perceptrons (MLPs), progressively leading to the final solution. These pivotal heads/MLPs, though identified on a specific dataset, exhibit transferability across different datasets and even distinct tasks.
This insight prompted us to investigate the potential benefits of selectively fine-tuning these essential heads/MLPs to boost the LLMs' computational performance. We empirically find that such precise tuning can yield notable enhancements on mathematical prowess, without compromising the performance on non-mathematical tasks. Our work serves as a preliminary exploration into the arithmetic calculation abilities inherent in LLMs, laying a solid foundation to reveal more intricate mathematical tasks.

%% file: tex/1_intro.tex
Large language models (LLMs) have experienced rapid advancements and shown impressive language understanding capabilities \citep{bert, few-shot, palm}. Notably, LLMs exhibit emergent abilities \citep{emergent} that enable them to solve intricate reasoning tasks akin to humans, such as mathematical computations \citep{chatgpt-math, math2}, chain-of-thought reasoning \citep{cot, cot-2}, few-shot prompting \citep{few-shot, flamingo}, etc. Despite these impressive characteristics, the complex inner processes governing LLMs' functionality have yet to be fully illuminated, due to the complex and intricate non-linear interactions within densely-connected layers. Comprehending these underlying mechanisms could contribute to predicting how the LLMs behave beyond their training data \citep{mu2020compositionalExplanations}, gaining insights into the emergence of certain behaviors \citep{meca_interp_grokking,barak2022hidden,Wei2022EmergentAO}, as well as identifying and rectifying errors present in the specific models \citep{hernandez2021natural,vig2020investigating}.

\begin{figure}[htbp]
  \centering
    \includegraphics[width=1.0\linewidth]{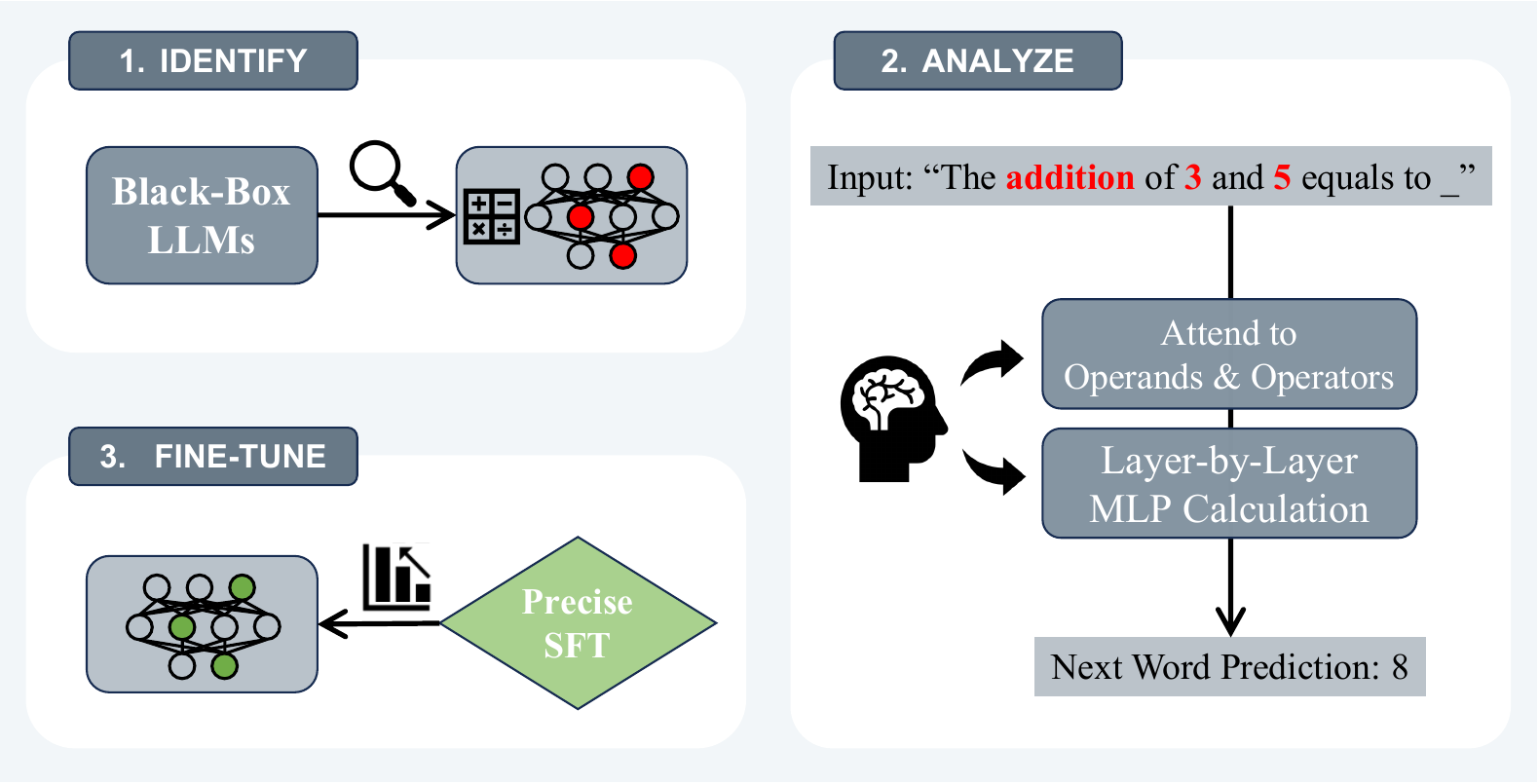}
    \caption{The pipeline involves three steps: 1) \textit{identify} the key components attributed to arithmetic calculations in black-box LLMs, 2) \textit{analyze} the working mechanism of the key components towards human-understandable explanations, 3) \textit{fine-tune} the key components to precisely improve the mathematical capability of LLMs.
    }
  \vspace{-3mm}
  \label{fig:pipeline}
\end{figure}

In this work, we take the first attempt to interpret the inner process of LLMs through the lens of mathematical computation problems, which are conducted on publicly available LLMs (\eg, LLaMA2 series \cite{touvron2023llama}).
Unlike typical language comprehension tasks, 
mathematical computation tasks involve concise problem statements with definitive correct answers, requiring a process of reasoning and calculation rather than direct copying to derive the solutions. These characteristics enable us to gain insights into the models' reasoning capabilities without interference from unrelated factors. Specifically, we focus on tasks involving the arithmetic calculation with two operands, \ie, addition, subtraction, multiplication, and division, which are fundamentals of mathematical computation. To this end, we create datasets of various types of sentences that involve the calculation logic, such as ``The addition of 3 and 5 equals to $\_$'' in Figure~\ref{fig:pipeline}. 
The LLMs could provide answers with high confidence scores of over $80\%$ on average.

To unveil how these models correctly complete the task (\eg, ``$3+5=8$''), we begin by identifying the task-related internal components in LLMs. 
We do a hard intervention~\citep{pearl2009causality} on the transformer attention heads and multi-layer perceptrons (MLPs) to observe their effects on the predicted logits\footnote{Here, doing a hard intervention is equal to replacing the value of attention heads and MLPs, while performing a soft intervention means modifying the modules for calculating the attention and MLP values~\citep{pearl2009causality}.}. Our findings reveal that only a small percentage ($<5\%$) of the attention heads and the MLPs after these heads significantly impact the model's performance. Namely, LLMs frequently involve these attention heads and the subsequent MLPs when completing the calculations. 
Subsequently, we knock out these frequently-involved heads/MLPs to validate their faithfulness.
We find that the model performance decreases sharply when those pivotal heads/MLPs are knocked out, resulting in a decrease of around $70\%$ in accuracy.

To interpret the working mechanism of identified heads/MLPs towards human-understandable explanations, we gain a deeper analysis of their operational ``behaviors''. 
Specifically, we investigate the attention patterns of the crucial heads, and find that these attention heads exhibit a strong focus on the tokens representing operands and operators within mathematical sentences, demonstrating a relative insensitivity to other non-relevant tokens. For the analysis of MLPs, we compare the correlations between the embeddings of MLPs' input/output and the embeddings of number tokens (\ie, operands and answers). 
It reveals that the MLPs, guided by these number-attended heads, take operands as input, and mirror the attributes of tokens corresponding to correct answers more closely. 
These observations lead us to hypothesize that \textit{LLMs may initially employ a set of heads to pinpoint arithmetic operands from text, subsequently engaging MLPs to work out the answers.}
Additionally, the observed behaviors of these heads/MLPs exhibit a high degree of transferability, analogous to adversarial examples being \emph{transferable across models}~\citep{szegedy2014intriguing}. Namely, the key heads/MLPs identified on one dataset are also effective for other datasets. For instance, their impact is noticeable on the publicly available math datasets (\eg, SVAMP \citep{SVAMP}), as well as varied data formats involving multi-digit integers, rational numbers, etc.
This empirical observation underscores the crucial role of key heads/MLPs in mathematical calculations.

In addition to uncovering the internal mechanisms, we have devised an effective strategy that involves targeted fine-tuning of the specific attention heads and MLPs closely tied to mathematical computations, thereby enhancing the model's mathematical prowess. The experimental results are compelling: with fine-tuning as few as $32$ attention heads (with a total of $1024$ heads), we observe a remarkable improvement in the model's mathematical capabilities. This precise tuning methodology not only matches but can surpass the enhancements achieved through full-model fine-tuning. Moreover, this fine-grained strategy of adjustment has a distinct advantage—it leaves most of the model's parameters unchanged, avoiding the performance trade-offs in non-mathematical domains commonly observed with full-model fine-tuning.

In summary, this work aims to delve into the inner mechanism of LLMs through mathematical calculation tasks, along the pipeline of ``identify-analyze-finetune'' shown in Figure~\ref{fig:pipeline}. Our findings reveal a sparsity in the attention heads of LLMs, with less than $5\%$ of heads exhibiting close correlations. These heads particularly attend to the operands and operators, while the subsequent MLPs gradually deduce the correct answers. The discovered mechanism shows strong cross-dataset transferability and inspires us to precisely finetune the calculation-related heads/MLPs for better mathematical capability. We empirically find that precise tuning brings in much less impact on non-mathematical tasks when improving the targeted ability of LLMs.

%% file: tex/5_related_works.tex
\textbf{Interpretability Methods.}
Interpreting the inner mechanism of large language models (LLMs) has become increasingly urgent in recent years \citep{madsen2022post, rauker2023toward}, especially when LLMs are applied in high-stakes decision-making domains such as healthcare, criminal justice, and finance \citep{obermeyer2019dissecting, rudin2019stop, bender2021dangers}. 
\citet{vig2020investigating} adapted the approach of \textit{causal mediation analysis} (CMA) \citep{CMA-theory} for interpreting the deep language models, and it has been applied for various tasks, such as subject-verb agreement \citep{CMA-subj}, natural language inference \citep{CMA-natural}, retention of factual associations \citep{CMA-locating, CMA-factual}. 
Furthermore, \textit{path patching} extends the concept of CMA by measuring how a treatment effect is mediated by node-to-node connections between individual neurons or features. 
Recent works have used path patching to explain neural networks in terms of circuits \citep{circuit}, identified for different capabilities including indirect object identification \citep{inter-IOI}, greater-than computation \citep{greater-than}, and mapping answer text to answer labels \citep{pp-choice}.

\textbf{Interpretability for Mathematical Tasks.}
Mathematical ability has long been a subject of interest in natural language processing~\citep{math-interest-3, math-interest-4, WangLS17, ThawaniPIS21}. 
Some studies have investigated the mathematical abilities of LLMs \citep{chatgpt-math, math-problem-2, math-behavior-1, math-behavior-2, MathPrompter, MathProgram}, but they mainly focus on explaining \textit{what} these models can do rather than \textit{how} they do it. 
In contrast, some other studies have dived deeper into the LLM structure without treating LLM as an inscrutable black box.
\citet{arithmetic} identified the key attention \textit{layers} relating to arithmetic questions, but lacking in-depth explanation and validation of the key layers' behaviors. 
\citet{Alpaca-inter} scaled the methods from causal abstraction to understand how Alpaca ($7$B) \citep{Alpaca-model} follows the instruction in comparing two numbers.
\cite{greater-than} provided a causal explanation about how GPT2-small ($0.1$B) \citep{gpt2} implements the ``greater-than'' task, but only reveal simple phenomena limited by the small size of model and the lack of diversity in the dataset. 

\textbf{Fine-tune LLMs for Mathematical Tasks.}
Numerous studies improve the mathematical reasoning ability of LLMs by aggregating various sampled reasoning paths during either fine-tuning or inference. 
\citet{gsm8k} train and devise a reasoning path verifier to select the correct results during inference. 
\citet{wang2023selfconsistency} propose to sample various reasoning paths during inference and then derive the final result by majority voting on the answers or through verifiers \citep{li-etal-2023-making}. 
\citet{uesato2022solving} explore to use of reinforcement learning methods for improving the mathematical reasoning abilities of LLMs.
Several works apply the idea of rejection sampling along with other techniques to filter the diverse sampled reasoning paths for fine-tuning data augmentation \citep{huang2022large,zelikman2022star,ni2023learning}. There also exist related works \cite{graft} that locate key parameters to update for better task-specific ability. 
\citet{graft} locates a minuscule subset of parameters from an already fine-tuned model onto a pre-trained model without further tuning. The selection process for this subset is via optimizing the task-related objective function with L1 norm ensuring the sparsity of the subset. In our work, we locate the task-related parameters of pre-trained model via measuring the \textit{causal effect} of each component, then \textit{precisely fine-tune} the key components for mathematical tasks.

%% file: tex/2_background.tex
\textbf{Large Language Models (LLMs).}
The LLMs utilized in this work comprise LLaMA2-7B and LLaMA2-13B \citep{llama2}. These are pre-trained language models freely available from HuggingFace\footnote{https://huggingface.co/}. All of these models are decoder-only transformers equipped with multi-head attention (MHA) and a single MLP in one transformer layer. 
For example, LLaMA2-7B consists of $32$ transformer layers and $32$ attention heads in MHA for each layer.

\textbf{Transformer Architecture.} 
The input to the transformer is a combination of position and token embeddings in $\mathbb{R}^{N \times d}$, where $N$ is the number of tokens in the input and $d$ is the model dimension.
Following the definitions in \citep{elhage2021mathematical}, the input embedding serves as the initial value for the \textit{residual stream}, which is read from and written to by all attention heads and MLPs. 
Focusing on individual heads, the $j$-th head in the $i$-th layer is parametrized by four matrices: $W_Q^{i,j}$, $W_K^{i,j}$, $W_V^{i,j} \in \mathbb{R}^{d \times \frac{d}{H}}$, and $W_O^{i,j} \in \mathbb{R}^{\frac{d}{H} \times d}$.
To simplify these parameters, we can express them as low-rank matrices in $\mathbb{R}^{d \times d}$: $W_{OV}^{i,j} = W_O^{i,j} W_V^{i,j}$ and $W_{QK}^{i,j} = W_Q^{i,j} (W_K^{i,j})^T$. The QK matrix is used to compute the attention pattern $A_{i, j} \in \mathbb{R}^{N\times N}$ for head $(i, j)$, while the OV matrix determines the information written into the residual stream.
At the end of the forward pass, a layer norm is applied before the unembed matrix $W_U$ projects the residual stream into logits.

\textbf{Task and Dataset.} We focus on classic and widely encountered mathematical operations, \eg, addition, subtraction, multiplication, division. Taking addition as an example, the arithmetic logic of addition (\A\ + \B\ = \C) might naturally appear in sentences. 
Taking inspiration from the sentence styles and forms present in mathematical benchmarks of GSM8K \citep{gsm8k} and SVAMP \citep{SVAMP}, we create a dataset for the addition task containing $10,000$ samples based on $36$ templates with random single-token names, objects, and numbers. 
To assess the performance of LLMs on the calculation task, we measure the prediction probability of the \texttt{\{C\}} token. 
The average probability of correct predictions across the models was $82\%$.
In this study, we select the samples that the language models are able to predict correctly. We denote the sentences generated by this procedure as reference data using the notation of $\xr$. 
For the templates and sentences, please refer to Figure \ref{fig:base_templates} and Figure \ref{fig:table_templates} in Appendix~\ref{app:templates}.

Moreover, to meet the demand for perturbing component activation, we create another dataset comprising counterfactual sentences without the inclusion of calculation logic, using the notation of $\xc$. The samples are generated following two core principles: (1) maintaining the grammatical structures derived from the $\xr$ templates; (2) substituting several crucial words responsible for the calculation logic with irrelevant words. 
For example, the sentence from $\xr$ like ``42 plus 34 is equal to $\_$'' is replaced to the counterfactual one ``42 \textit{nothing} 34 is equal to $\_$''.
In this way, it allows for a direct reflection of the model's impact on the arithmetic calculation tasks, rather than being influenced by the sentence structure or syntax.

%% file: tex/3_method.tex
Our goal is to interpret the LLMs in a way that is human-understandable, thus enabling targeted modification of models through precise SFT. This section delves into the ``identify-analyze-finetune'' methodology.
First, in Section~\ref{sec:identify}, we describe the process of identifying and validating key components within LLMs. Then in Section~\ref{sec:analysis}, we examine the inherent patterns of these pivotal components to decode their behavior and distinct features. Finally, in Section~\ref{sec:finetune}, we introduce a strategy of precise SFT that fine-tunes these influential components to enhance the proficiency in calculation.

\subsection{Key Components Identification.}
\label{sec:identify}

The computation of the LLM can be reorganized as a directed acyclic graph (DAG)~\citep{inter-IOI}.
In the graph, each node is a computation component, including attention heads, MLP layers, residual connections, and each edge represents the data flow that the output of the previous node will be transposed to the input of the later node. Please refer to Appendix \ref{app:general_pp} for more details.
To unravel the underlying cause of the model's predicted answer, we employ the causal intervention technique known as \textit{path patching} \citep{localizing, inter-IOI}.
By perturbing targeted activation with counterfactual data $\xc$ and freezing others with reference data $\xr$, the comparison on output logits is employed to measure the counterfactual effect. The whole process is illustrated in Algorithm~\ref{alg:identify}.
In this work, we scan through all nodes $\mathcal{N}$ one by one, and measure the changes in the output logit of ground-truth token \C, recoding in $E_{\mathcal{N}}$. Notably, since the residual operations and MLPs compute each token separately \citep{elhage2021mathematical}, patching the head output at the END position (\ie, the last token in the input sentence) is enough to measure the effects on the next token prediction.

\begin{algorithm}[h]
\caption{Identifying Key Components}\label{alg:identify}
\begin{algorithmic}
\State \textbf{Input}: Set $\mathbf{\Omega}$ of reference and counterfactual sample pairs ($X_{\text{r}}$, $X_{\text{c}}$), model $\mathcal{M}$ with nodes $\mathcal{N}$.
\State \textbf{Output}: Causal effects for $\mathcal{N}$: $E_{\mathcal{N}}$.
   \For{$(X_r^{(i)}$, $X_c^{(i)})$ in $\mathbf{\Omega}$}
       \State Compute all activations $A_r$, $A_c$ on $(X_r^{(i)}$, $X_c^{(i)})$
       \For{n in $\mathcal{N}$}
           \State $A'_r(n) \gets A_c(n); $ \Comment{replace output in $A_r$ by $A_c$}
           \State $A'_r(k) \gets A_r(k), \forall k \in [1,\cdots,|\mathcal{N}|], k\neq n.$
           \State $logit_o \gets \mathcal{M}(X_r^{(i)}, A_r)$ \Comment{get original logits}
           \State $logit_p \gets \mathcal{M}(X_r^{(i)}, A'_r)$ \Comment{get patched logits}
           \State ${s}_{n}^{(i)} \gets \frac{logit_p - logit_o}{logit_o}$ \Comment{causal effect}
        \EndFor
   \EndFor
   \State \textbf{Return:} $\overline{s_n} = \frac{\sum_{i=1}^{|\mathbf{\Omega}|} s_n^{(i)}}{|\mathbf{\Omega}|}$ \Comment{averaged effect w.r.t. samples}

\end{algorithmic}
\end{algorithm}
Explanations for model behavior can easily be misleading or non-rigorous \citep{rigorous, rigorous-2}. To address this issue, we further assess the importance of the identified heads/MLPs, while also confirming the insignificance of others.
For this purpose, we employ a knockout technique called \textit{mean ablation} \citep{inter-IOI} to deactivate the individual heads/MLPs and observe their impact on model performance. Specifically, we replace their activation with average activation across counterfactual data $\xc$ to remove the task-related information. By observing changes in model performance, we can verify the roles of these key heads/MLPs.

\subsection{Pattern Analysis.}
\label{sec:analysis}
To make the identified heads/MLPs accessible to human understanding, we conduct a deeper analysis of their operational “behaviors”. For attention heads, we examine the attention pattern $A_{i,j}  \in \mathbb{R}^{N \times N}$ to comprehend which tokens are prioritized. $N$ is the number of input tokens. Specifically, we begin by gathering the respective attention patterns $A_{i,j}$ on reference data $X_r$ of the key heads. We extract the last row of $A_{i,j}$ for each sample, analyzing the attention scores $A_{i,j}^{END} \in \mathbb{R}^{1 \times N}$ between the Query token at the END position and each Key token, and obtaining the averaged scores w.r.t. samples.
Generally, the type of token with the highest attention score represents the characteristics of the head, such as numbers, math symbols, etc.

For MLPs, we use the unembedding matrix as the probing to measure the content of token, especially numerical tokens, contained in MLPs' inputs and outputs. 
Prior studies, such as those reported in \citep{elhage2021mathematical}, have illustrated that the MLP layer initially receives its input from the residual stream (\ie, $MLP_{in}$), subsequently adding its output back into that stream (\ie, $MLP_{out}$).
Let $W_U$ represent the unembedding matrix, and $W_U[*]$ denote the unembedding vector corresponding to a specific token. We calculate the cosine similarity between $MLP_{in}$, $MLP_{out}$ and $W_U[\mathrm{\texttt{\{A\}}}]$, $W_U[\mathrm{\texttt{\{B\}}}]$, $W_U[\mathrm{\texttt{\{C\}}}]$ to reflect the information the MLP receives and generates. To isolate the specific contribution of MLP to specific numerical tokens, we further evaluate the subtraction of outputs and inputs of MLP, \ie, $\frac{MLP_{out} - MLP_{in}}{||MLP_{out} - MLP_{in}||}\cdot \frac{W_U[\mathrm{\texttt{\{A\}}}]}{||W_U[\mathrm{\texttt{\{A\}}}]||}$.
Research in \citep{trans_sub_space} presents that each MLP layer’s output token representation can be characterized as an additive update influencing the evolving representation across vocabularies.
Our methodology is aligned with these works, while we mainly focus on the token embeddings of right/wrong answers to reveal the contribution of MLPs on the calculation tasks.

\subsection{Precise Fine-tuning.}
\label{sec:finetune}

Supervised Fine-Tuning (SFT) is widely used for enhancing a model's mathematical capabilities. Building on this, precise SFT only updates those components closely associated with mathematical abilities, while keeping the rest parameters unchanged. Algorithm~\ref{alg:precise_tune} illustrates the whole process.
For the $i$-th attention layer, the output matrix $W_O^{i}$ is split into equal size blocks for each head $\left [  W_O^{i,1}, W_O^{i,2}, \cdots W_O^{i,H}\right ]$. 
\begin{algorithm}[htbp]
\caption{Precise Fine-tuning}\label{alg:precise_tune}
\begin{algorithmic}
\State \textbf{Require}: Model $\mathcal{M}$, input $X$, index of key heads $\mathbf{\Phi}$, iterations $I$, learning rate $\eta$, $W_{\theta}=W_{Q/K/V/O}$
\State \textbf{for} $(i,j) \in \mathbf{\Phi}$ \textbf{do}
\State \qquad $W_{\theta}^{i,j}.requires\_grad = True$
\State \textbf{end for}\Comment{activate key heads}
\State \textbf{loop} $I$ times
\State \qquad $\mathcal{L}$ = $\mathcal{M}.$forward($X$)
\State \qquad $\mathcal{L}$.backward()
\State \qquad \textbf{for} $w\in W_{\theta}$ \textbf{do}
\State \qquad \qquad $w=w - \eta * w.grad$
\State \qquad \textbf{end for}\Comment{update target parameters}
\State \textbf{end loop}
\end{algorithmic}
\end{algorithm}
As is verified in \citep{elhage2021mathematical}, it is equivalent to running heads independently, multiplying each by its own output matrix, and adding them into the residual stream. 
For the selected individual heads, precise SFT updates the parameters of four matrices: $W_Q^{i,j}$, $W_K^{i,j}$, $W_V^{i,j} \in \mathbb{R}^{d \times \frac{d}{H}}$, and $W_O^{i,j} \in \mathbb{R}^{\frac{d}{H} \times d}$. For the selected MLP layer, precise SFT updates all parameters in this layer.
Moreover, since we adjust only a small fraction of the parameters, precise SFT naturally benefits from shorter training times and minimal impact on the model's original capabilities.

%% file: tex/4_exp.tex
The experiments are organized as follows: (1) \textit{identify} the calculation-related key components via path patching and \textit{validate} their importance in implementing arithmetic calculation via knockout in Section~\ref{exp:discover}; (2) \textit{understand} the behavior of the newly identified components by examining their attention patterns and embeddings in Section~\ref{exp:understand}; 
(3) \textit{improve} the mathematical capability via precise supervised fine-tuning on math benchmarks in Section~\ref{exp:finetune}.
For simplicity, we primarily report the results of LLaMA2-7B, while the results of other models can be found in Appendix.

\begin{figure}[htbp]
  \centering
    \includegraphics[width=0.96\linewidth]{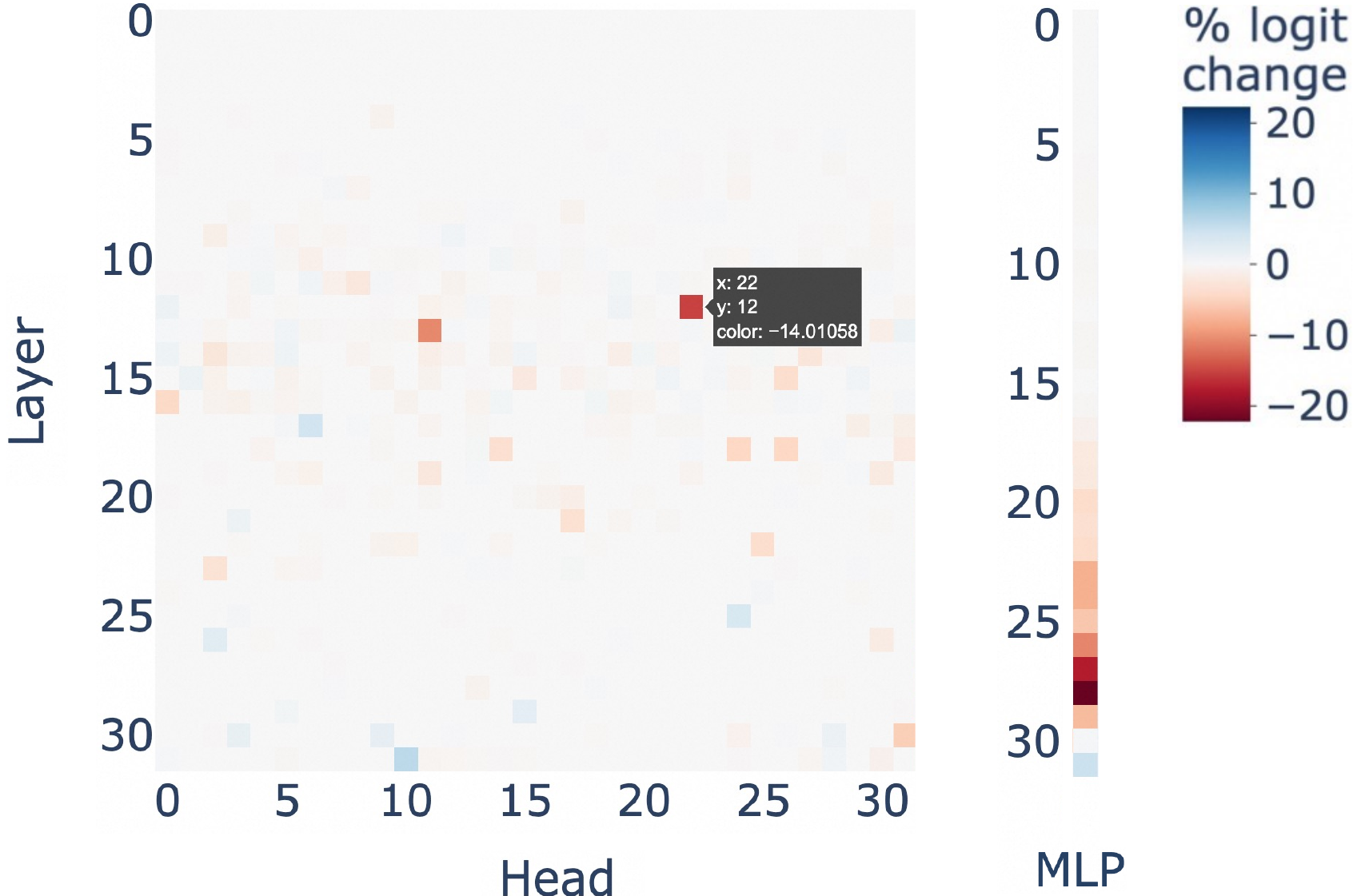}
    \caption{We conduct path patching experiments on LLaMA2-7B across four mathematical tasks, by searching for each head and MLP directly affecting the logit of the right answer. For each head/MLP, a darker color indicates a larger logit difference from the model before patching. }
  \label{fig:app_four_tasks}
    \vspace{-2mm}
\end{figure}

\subsection{Identifying Calculation-related Components.
}
\label{exp:discover}

\textbf{Location of key heads.}
In Figure \ref{fig:app_four_tasks}, we visualize the effect of each head according to the serial numbers of the heads and layers. This arrangement allows for a clear comparison of the causal impact of each head to the logit of ground-truth token \C. The red squares indicate heads that have a significant positive impact on predicting the output token, while the blue squares represent heads that have a negative effect. From these results, we observe that: 
(i) \textit{Only a small number of heads have a noteworthy influence on the output}. Specifically,  when the heads such as $12.22$\footnote{We apply the notation of $i.j$ to refer to the $j$-th head of the $i$-th attention layer.} is patched, there is a substantial decrease of $14.0\%$ on the logit of token \C, 
which highlights their positive contribution to the calculation tasks. We classify heads that exhibit logit change exceeding $-5\%$ as ``key heads''.
The sparse distribution of these key heads motivates us to explore their specific functionalities and characteristics in Section~\ref{exp:understand}. 
(ii) \textit{The discovered key heads are mainly located in the middle layers}. For LLaMA2-7B, key heads emerge starting from the $12$th layer for all arithmetic calculations. Prior layers exhibit heads that do not exert a direct effect on the output logits. Key heads are primarily concentrated between layers $12$ and $17$.
(More analysis of the key heads in other LLMs can be found in Appendix~\ref{app:PathP_LLM}.)

\textbf{Location of key MLPs.}
The last column in Figure~\ref{fig:app_four_tasks} visualizes the effect of each MLP layer on the logit of ground-truth token \C. It is observed that MLPs before the identified heads ($0-16$) have almost no impact on the outputs (approximately $\pm 0.0\%$). In contrast, after the $17$-th layer, MLPs exhibit a much larger effect (approximately $\pm 10.0\%$).
It indicates that MLPs are engaged in the calculation. 
We hypothesize that the calculation process is firstly implemented through the key heads, then the subsequent MLPs gradually work out the final results. We validate this in Section~\ref{exp:understand}.

\begin{figure}[tbp]
  \centering
    \includegraphics[width=0.9\linewidth]{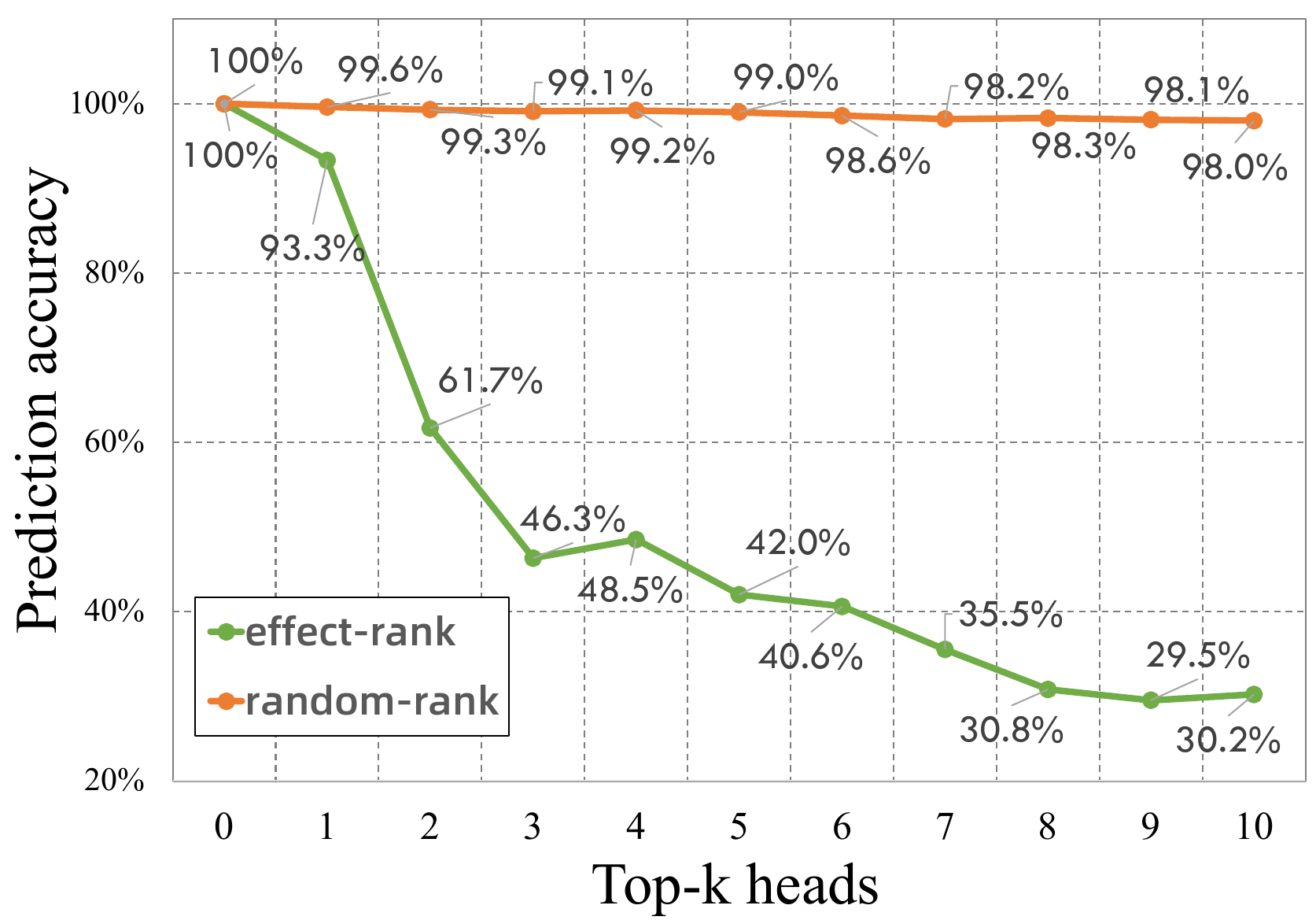}
    \caption{The influence on prediction accuracy after knocking out top-k attention heads that are sorted by the effect of each head on logits (``effect-rank''), and knocking out randomly-sorted top-k heads (``random-rank'').}
  \label{fig:valiate}
\end{figure}

\begin{figure}[tbp]
  \centering
    \includegraphics[width=\linewidth]{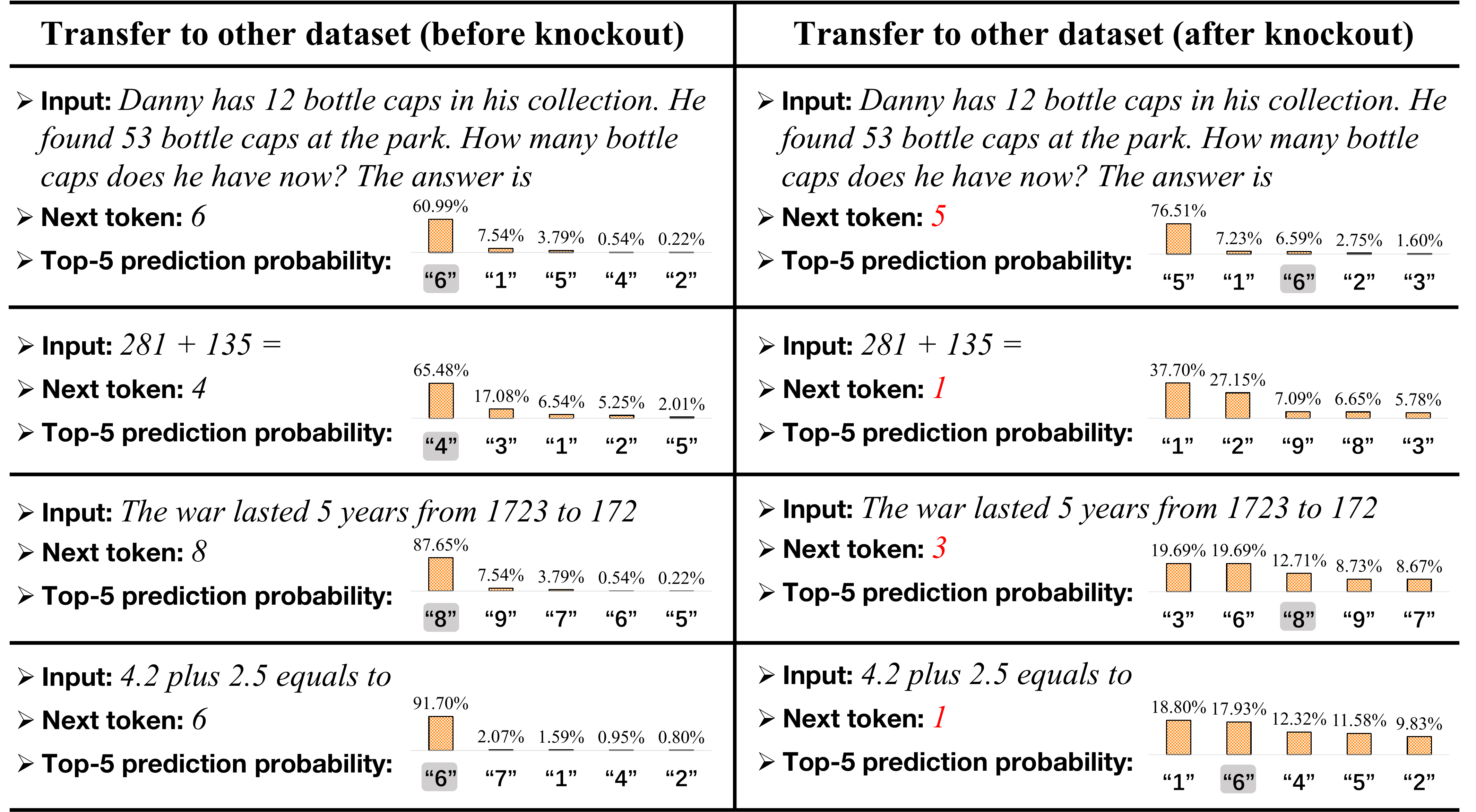}
    \caption{After knocking out the key heads, LLaMA2-7B predicts incorrectly on the cases of SVAMP dataset and other data formats of multi-digit integers, rational numbers.}
  \label{fig:knock_cases}
\end{figure}

\textbf{Validation of key components.}
To fully validate the faithfulness of the discovered key heads, we perform additional checks by observing the performance drop when knocking out these components. 
In Figure~\ref{fig:valiate}, all heads are sorted in a certain order by the importance score shown in Fig. \ref{fig:app_four_tasks} and knocked out one by one. 
It shows that, as the heads are gradually knocked out, the performance of the model drops sharply in ``effect-rank'', while keeping stable (relatively minor effect within $2\%$) in ``random-rank''.
We also exhibit the transferability of the key heads with different data prompts or formats as shown in Figure~\ref{fig:knock_cases}. The model becomes largely confused to output incorrect numbers after knocking out the identified key heads.
On the dataset SVAMP, there is a relative performance drop ($-22.9\%$/$34.7\%$=$-66.0\%$) after the knockout, aligned with the result on our generated dataset.
The above results demonstrate that the discovered components play an important role in the language model's ability to complete the calculation task.

\begin{figure}[tbp]
  \centering
\subfigure[Addition]{
    \includegraphics[width=0.42\linewidth]{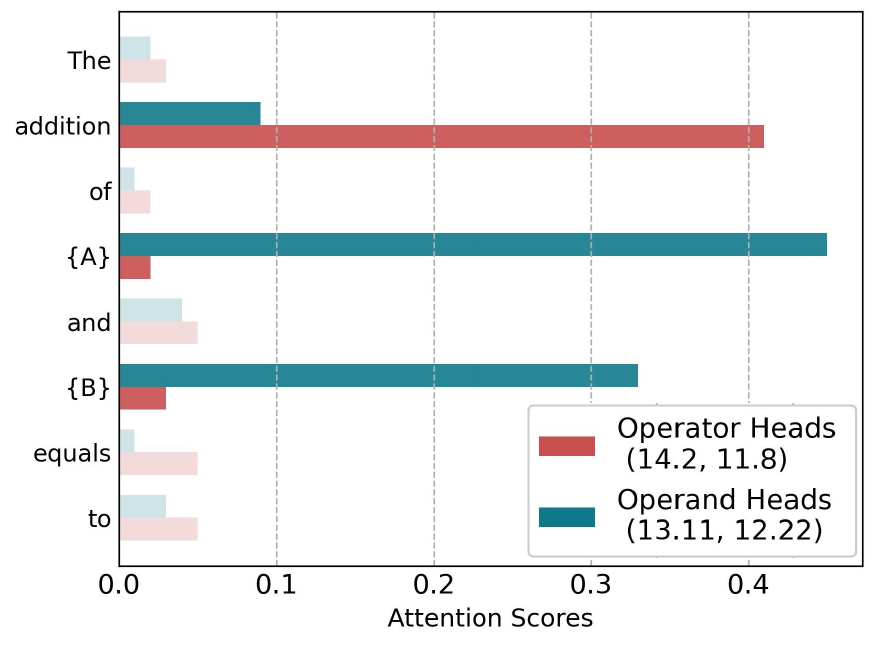}
  \label{fig:attn_analysis_add} 
  }
  \hfill
  \subfigure[Subtraction]{
    \includegraphics[width=0.42\linewidth]{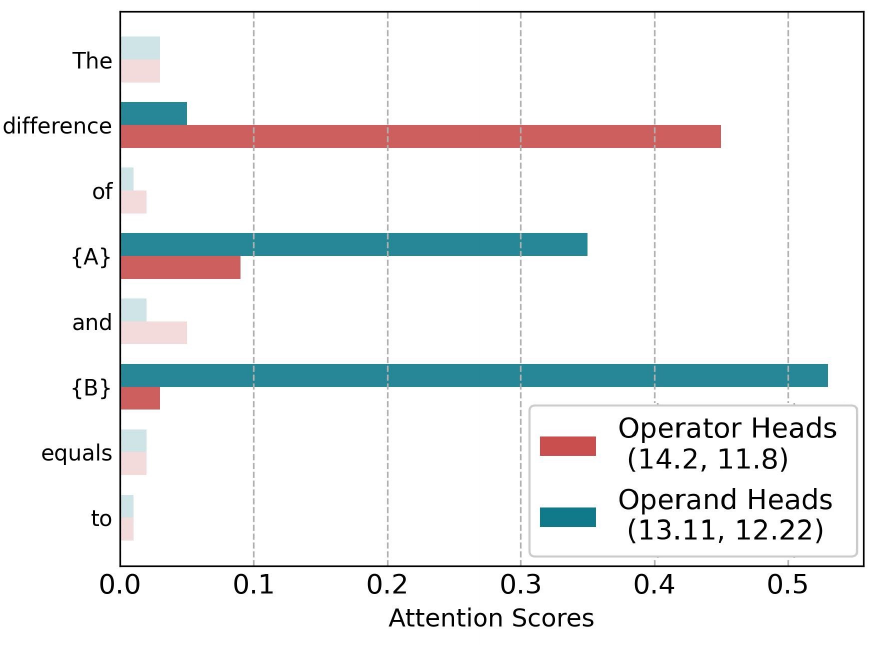}
  \label{fig:attn_analysis_sub} 
  }
  \\
  \subfigure[Multiplication]{
    \includegraphics[width=0.42\linewidth]{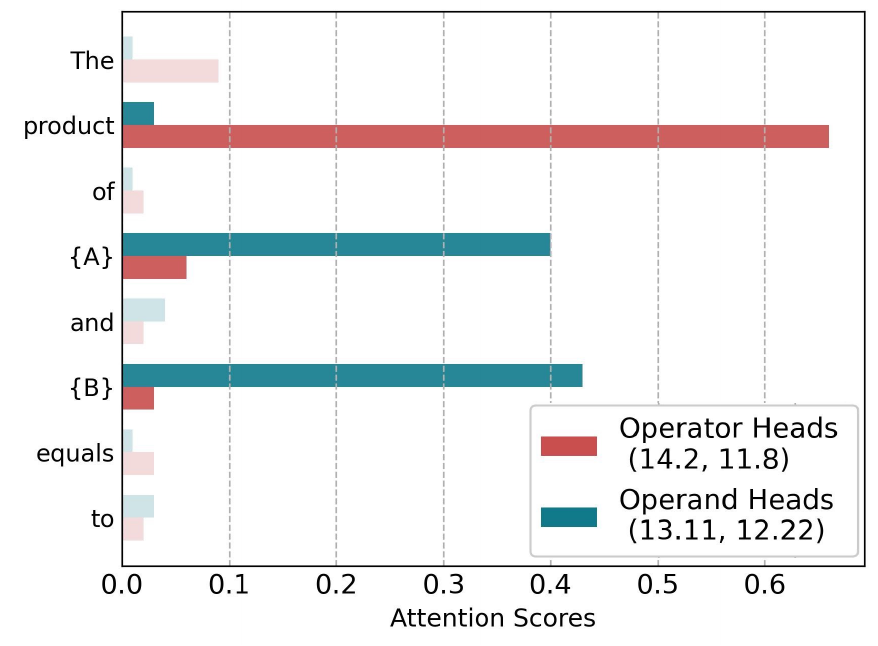}
  \label{fig:attn_analysis_mul} 
  }
  \hfill
  \subfigure[Division]{
    \includegraphics[width=0.42\linewidth]{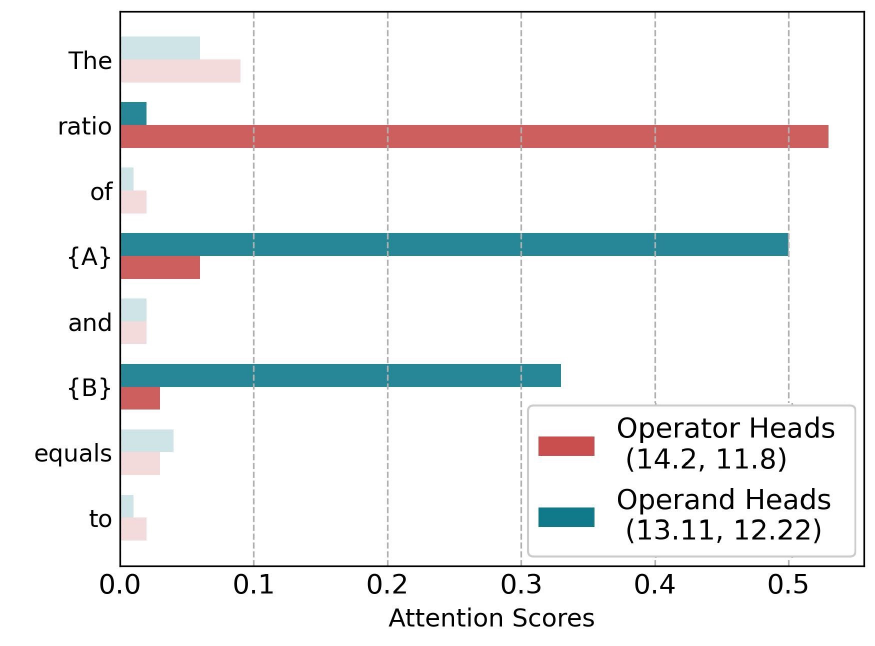}
  \label{fig:attn_analysis_div} 
  }
  \caption{The attention score distribution of key heads across four calculation tasks. The key heads (\eg, $13.11$, $14.2$) attend to number operands and calculation operators.}
   \label{fig:attn_analysis} 
\end{figure}

\subsection{Understanding Calculation-related Component Behaviors.}
\label{exp:understand}

\textbf{Key heads behavior.}
In order to better understand the ``behavior'' of the heads that have a significant impact on calculation, we begin by analyzing their attention patterns, and check the attention scores between Query END token and each Key token as illustrated in Sec. \ref{sec:analysis}. 
Our findings reveal that these heads exhibit a strong focus on tokens of operands or operators. For example, heads $13.11$ and $12.22$ have high attention scores on numbers including \A\ and \B, while heads $14.2$ and $11.8$ attend more to symbols or text indicating operations like ``+'', ``-'', ``plus'', ``div'', etc. We randomly select $1000$ samples from reference data and plot the distribution of averaged attention scores on key heads (arranged in two groups) for four arithmetic calculations.

\begin{figure*}[!t]
  \centering
\subfigure[Reception of \A\ and \B]{
    \includegraphics[width=0.28\textwidth]{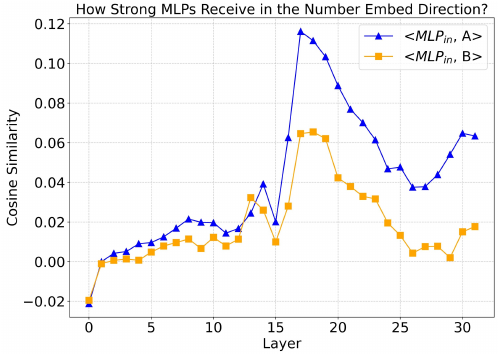}
  \label{fig:MLP_in} 
  }
  \subfigure[Generation of \C]{
    \includegraphics[width=0.28\textwidth]{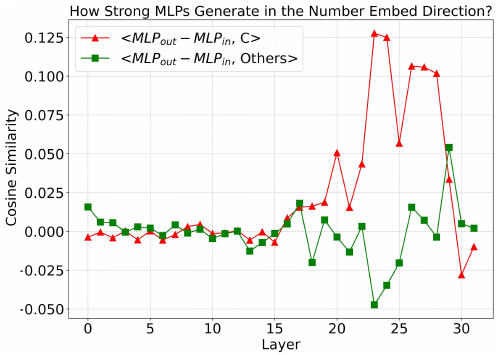}
    \label{fig:MLP_diff} 
  }
  \subfigure[Steps of Calculation]{
    \includegraphics[width=0.38\textwidth]{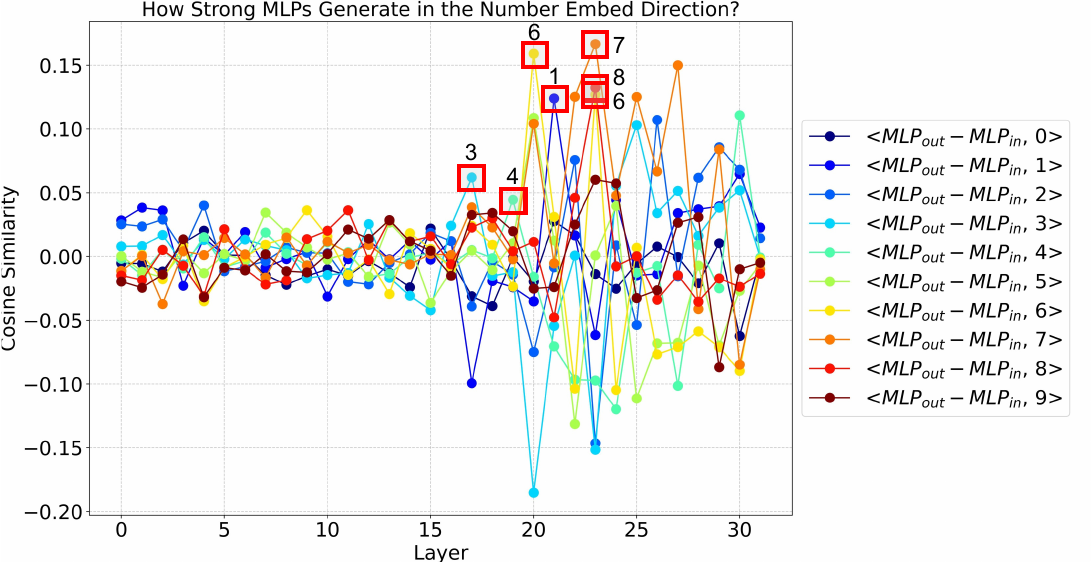}
    \label{fig:MLP_diff_3+4} 
  }
  \caption{We investigate the projection of each MLP layer input or output along the direction of number token \A, \B, and \C, respectively. The x-axis represents the layer number, ranging from 0 to 31, while the y-axis represents the cosine similarity between the embeddings of the MLP input or output and the number tokens.}
   \label{fig:result_scatter_mlp} 
\end{figure*}

\begin{figure}[htbp]
  \centering
\subfigure[SVAMP]{
    \includegraphics[width=0.37\linewidth]{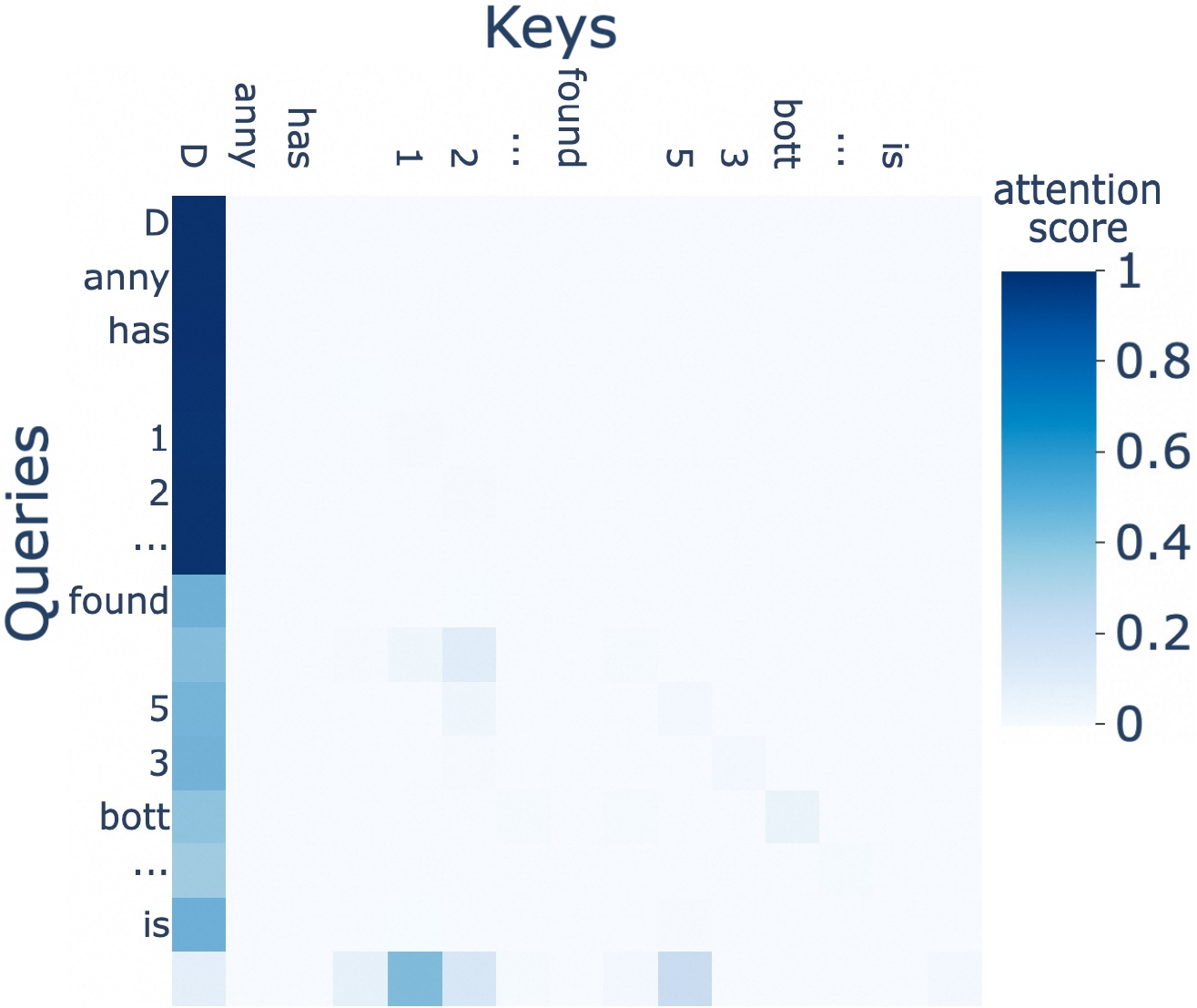}
  }
  \hskip 1.5em
  \subfigure[Multi-digit integer]{
    \includegraphics[width=0.37\linewidth]{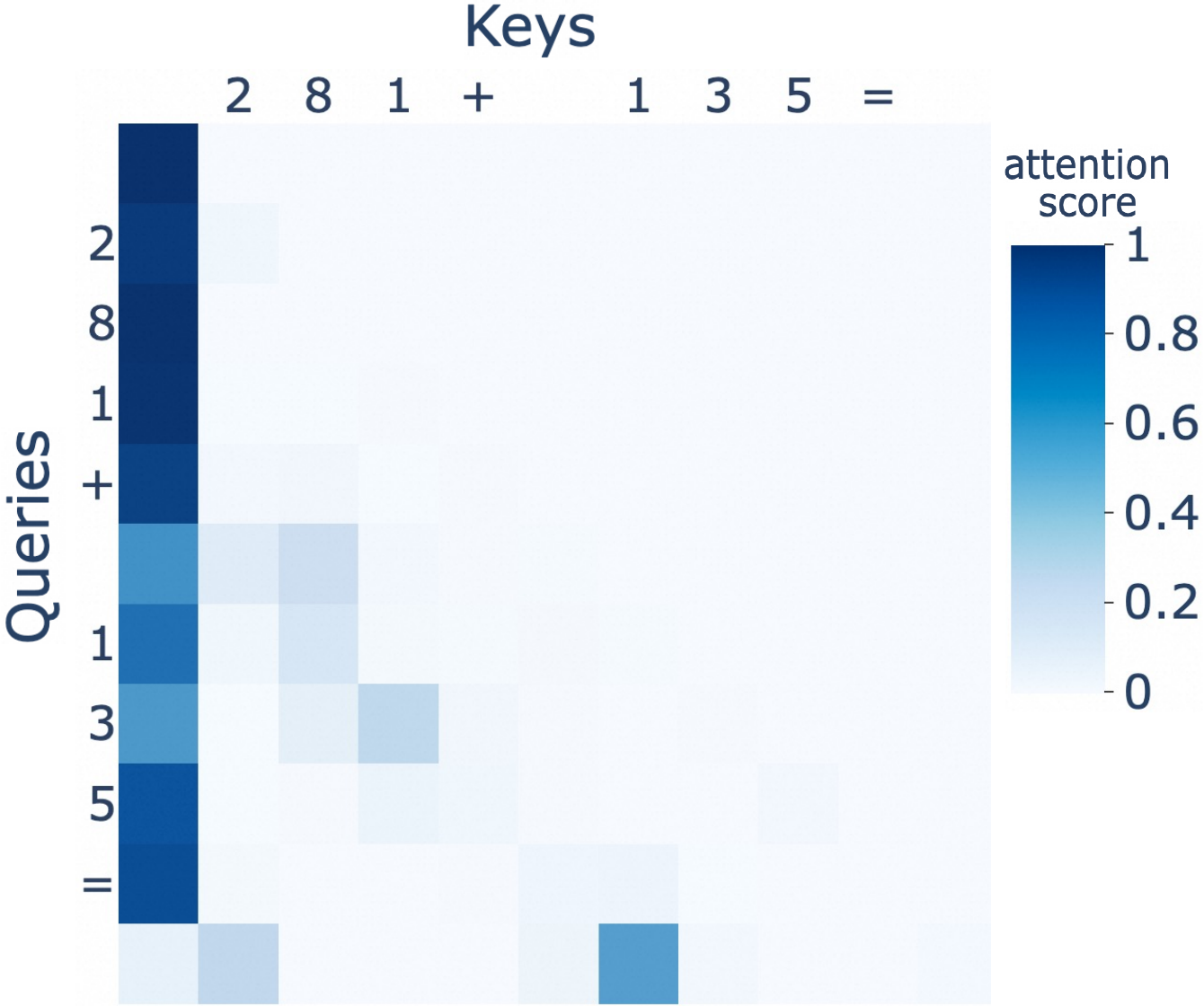}
  }
  \\
  \subfigure[Diff. data format]{
    \includegraphics[width=0.37\linewidth]{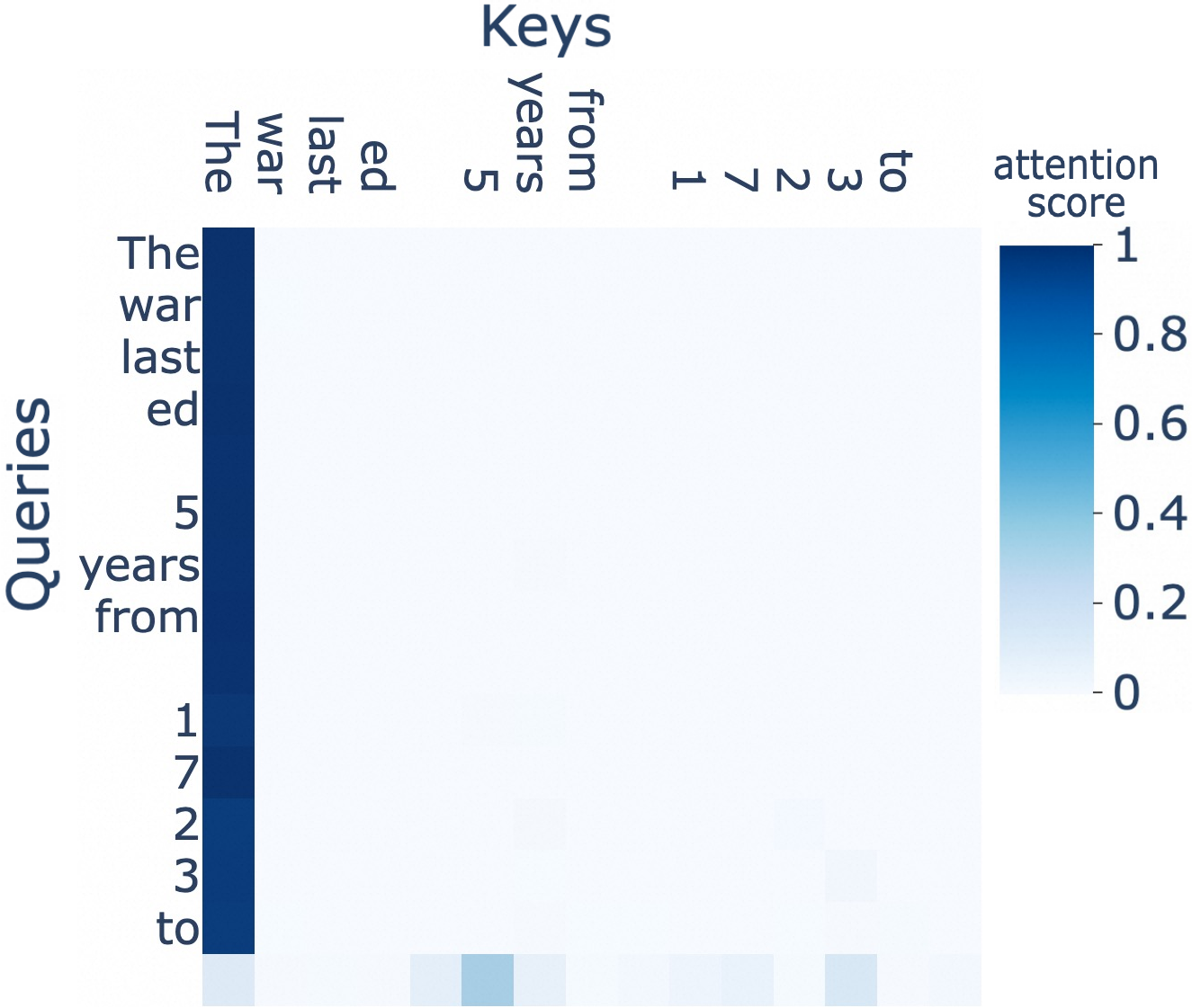}
  }
  \hskip 1.5em
  \subfigure[Rational numbers]{
    \includegraphics[width=0.37\linewidth]{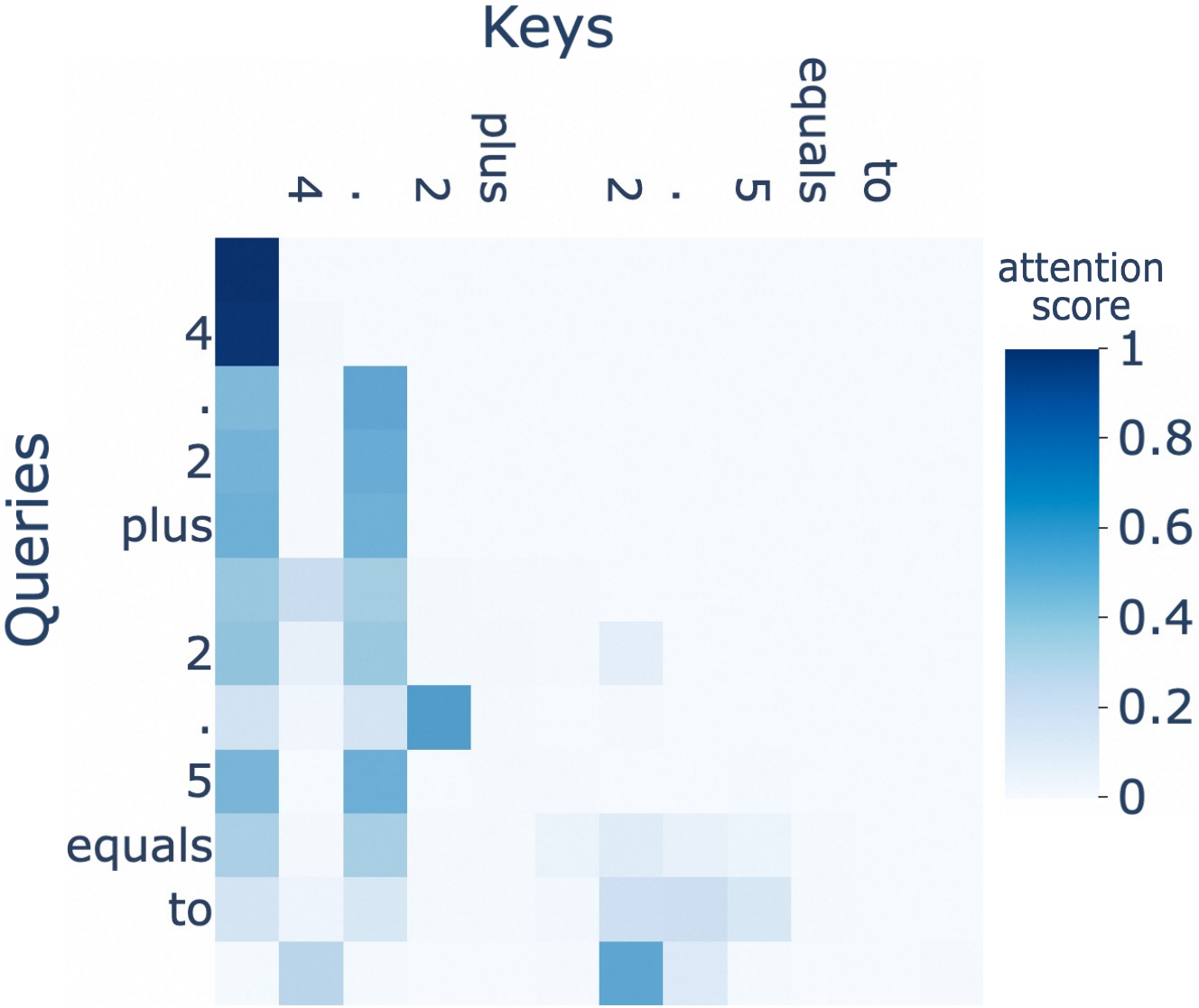}
  }
  \caption{The transferability of attention patterns in key heads on the unseen samples in Figure \ref{fig:knock_cases}, which mainly attend to the number operands.}
   \label{fig:attn_analysis_case} 
\end{figure}

As illustrated in Figure~\ref{fig:attn_analysis}, the operand heads and the operator heads are colored in red and green respectively, and highlighted at the positions of operands and operators. It is clear that these heads exhibit distinctly different distributions and show minimal attention to tokens outside of the operands/operators.
Moreover, we visualize the attention patterns of the key heads (\eg, $13.11$) on various types of sentences in Figure~\ref{fig:attn_analysis_case}.
It reveals that the key heads also primarily prioritize number operands (\eg, `$1$' and `$5$' in the first case) even for unseen data formats. This observation provides an explanation for why the deactivation of the key heads can influence the model's perception on number tokens and consequently affect its prediction when transferring to other datasets (shown in Figure~\ref{fig:knock_cases}).
For more case studies on the key heads, such as the attention pattern on operators, please refer to Figure \ref{app:attn_operator} in Appendix~\ref{app:attn_pattern}.

\textbf{Key MLPs behavior.}
In Figure~\ref{fig:MLP_in}, we conduct an initial investigation of the similarities between the $MLP_{in}$ and tokens \A\ and \B\ over $1000$ samples, to verify the information of operands received from above analyzed attention heads.
For the $0$--$12$th layers, both $\langle MLP_{in}, \A \rangle $ and $\langle MLP_{in}, \B \rangle $ are close to zero. It indicates no operands are captured during this stage, which corresponds to the blank region (\ie, few key heads for computation task) before the $12$th layer in Figure~\ref{fig:app_four_tasks}.
For the $12$--$17$th layers, we observe a sharp increase in the similarities with both operands (\A\ and \B). This surge corresponds to the presence of key attention heads, \eg, $12.22$/$13.11$ in layer $12/13$, indicating that the operands are progressively being collected and ``written'' into the MLPs of these layers for subsequent computations.
In layers $17$--$31$, the similarities $\langle MLP_{in}, \A\rangle$ and $\langle MLP_{in}, \B \rangle $ gradually decrease, signifying the transition into a new stage that digests the input information for generating the answers.

To understand how each MLP layer contributes to generating the correct answer \C, we compute the similarity between token \C\ and the input/output of the MLPs. We use $\langle MLP_{out} - MLP_{in}, W_U[\C]\rangle $ to reflect the direct contribution of the MLP to the correct answer, and $\langle MLP_{out} - MLP_{in}, W_U[Other] \rangle $ for other candidate numbers (as shown in Figure~\ref{fig:MLP_diff}). 
Starting from the $17$th layer, where the MLPs begin processing operand information, we observe a noticeable increase in $\langle MLP_{out} - MLP_{in}, W_U[\C] \rangle $ and a decrease in $\langle MLP_{out} - MLP_{in}, W_U[Other] \rangle $. 
This trend indicates that these MLPs are gradually carrying out the calculation required for the correct answer.
The above ascending and descending trends can also be viewed in other LLMs as in Figure \ref{app:other_model_receive} and Figure \ref{app:other_model_generate} in Appendix \ref{app:PathP_LLM}.

\newcommand{\sepsmall}[0]{\hspace{4pt}}
\newcommand{\septiny}[0]{\hspace{8pt}}
\begin{table*}[!t]
  \caption{
Overall performance. We evaluate the capabilities of LLaMA2-7B and LLaMA2-13B, transitioning from generic tasks (\eg, MMLU and CSQA) to mathematical tasks (\eg, GSM8K, AddSub, SingleEq, and SVAMP). Supervised fine-tuning across the entire parameter set (denoted as Full SFT) leads to enhanced performance in math-related tasks, albeit at the expense of its capabilities in generic tasks. In contrast, selectively tuning only the parameters of $32$ critical attention heads (denoted as Precise SFT) yields comparable improvements while preserving the model's proficiency in generic tasks, with faster training speed (samples processed per second) and less tuned parameters.
}
  \label{tab:exp:overall}
  \centering
  \setlength{\tabcolsep}{4pt}
  \begin{tabular}{
            @{}l @{}c @{}c  @{\hskip 10pt}
            c@{\sepsmall}c  m{0.01em} 
            c@{\sepsmall}c  m{0.01em} 
            c@{\sepsmall}c  m{0.01em} 
            c@{\sepsmall}c  m{0.01em} 
            @{\hskip 6pt}
            c@{\sepsmall}c  m{0.01em} 
            c@{\sepsmall}c 
            @{}}
      \toprule[1.25pt]
       &&&  \multicolumn{11}{c}{\textbf{Mathematical Tasks}} && \multicolumn{5}{c}{\textbf{Generic Tasks}} \\ 
       
       \cmidrule(lr){4-14}
       \cmidrule(lr){16-20}
        & & & \multicolumn{2}{c}{\textbf{GSM8K}} && \multicolumn{2}{c}{\textbf{AddSub}} && \multicolumn{2}{c}{\textbf{SingleEq}} && \multicolumn{2}{c}{\textbf{SVAMP}} 
        && \multicolumn{2}{c}{\textbf{MMLU}} && \multicolumn{2}{c}{\textbf{CSQA}} \\
        
       \cmidrule(lr){4-5}
       \cmidrule(lr){7-8}
       \cmidrule(lr){10-11}
       \cmidrule(lr){13-14}
       \cmidrule(lr){16-17}
       \cmidrule(lr){19-20}
        \textbf{Models}
        & \makecell[c]{\textbf{Train} \\ \textbf{Speed}} & \makecell[c]{\textbf{Tuned} \\ \textbf{Params.}} 
        &   \textbf{Acc.} & $\mathbf{\Delta}$
        &&  \textbf{Acc.} & $\mathbf{\Delta}$
        &&  \textbf{Acc.} & $\mathbf{\Delta}$
        &&  \textbf{Acc.} & $\mathbf{\Delta}$
        &&  \textbf{Acc.} & $\mathbf{\Delta}$
        &&  \textbf{Acc.} & $\mathbf{\Delta}$\\
        
      \midrule[1.25pt]
      
      LLaMA2-7B
      & -
      & -
      &  14.6 & - 
      && 30.5 & - 
      && 65.4 & - 
      && 34.7 & - 
      && 46.0 & - 
      && 59.8 & - \\

      \; + Full SFT 
      &  \; 15sam./sec.  
      &  6.7B  
      &  24.6  & \smallgreen{+10.0} 
      && 53.7 & \smallgreen{+23.2} 
      && 68.2 & \smallgreen{+2.8} 
      && 50.3 & \smallgreen{+15.6} 
      && 40.5 & \smallred{-5.5} 
      && 54.0  & \smallred{-5.8} \\ 

      \; + Precise SFT 
      &  \; 50sam./sec. 
      &  0.07B  
      &  27.4  & \smallgreen{+12.8} 
      && 50.6 & \smallgreen{+20.1} 
      && 69.7 & \smallgreen{+4.3} 
      && 55.8 & \smallgreen{+21.1} 
      && 46.4 & \smallgreen{+0.4} 
      && 59.6  & \smallred{-0.2} \\ 
      
      \midrule[0.5pt]
      
      LLaMA2-13B
      & -
      & -
      &  28.7 & - 
      && 33.7 & - 
      && 76.6 & - 
      && 45.7 & - 
      && 54.8 & - 
      && 67.3 & - \\

      \; + Full SFT 
      &  \; 8sam./sec.  
      &  13.0B  
      &  44.6  & \smallgreen{+15.9} 
      && 62.2 & \smallgreen{+28.5} 
      && 79.8 & \smallgreen{+3.2} 
      && 62.8 & \smallgreen{+17.1} 
      && 50.2 & \smallred{-4.6} 
      && 62.0  & \smallred{-5.3} \\ 

      \; + Precise SFT 
      &  \; 34sam./sec. 
      &  0.08B  
      &  46.3  & \smallgreen{+17.6} 
      && 61.1 & \smallgreen{+27.4} 
      && 82.2 & \smallgreen{+5.6} 
      && 66.6 & \smallgreen{+20.9} 
      && 55.0 & \smallgreen{+0.2} 
      && 67.2  & \smallred{-0.1} \\

      \bottomrule[1.25pt]
  \end{tabular}
\end{table*}

Based on the above analyses, we further delve into the detailed calculation process from layer $17$ to $28$.
We investigate a case of ``$4 + 3 = $ '' and analyze $MLP_{out} - MLP_{in}$ compared to all numeric tokens in Figure~\ref{fig:MLP_diff_3+4}.
At layers $17$ and $19$, the numbers `$3$' and `$4$' are at the top, indicating that MLPs receive and store input $\A=$`$4$' and $\B=$`$3$', respectively. After that, the numbers `$6$' and `$1$' appear top at the subsequent layers $20$ and $21$.
In summary, the LLM predicts the next token as `$7$' in a single inference. However, within the LLM's architecture, the answer `$7$' is the result of a collaborative process across multiple layers $22$/$23$/$25$/$27$, after the layers $17$/$19$/$20$/$21$ generate `$3$'/`$4$'/`$6$'/`$1$', respectively.
The results demonstrate that the answer `$7$' is not deduced directly, and MLPs perform calculations in a ``layer-by-layer'' manner, somewhat akin to the addition process in computers (a comparison of these two processes are presented Appendix~\ref{app:computer_vs_llms}). 
Additionally, we observe that numbers close to the correct answer, such as `$6$' and `$8$', also appear at the top in layer $23$.
However, in subsequent layers, the correct answer `$7$' consistently remains top while `$6$' and `$8$' decline. It indicates that LLMs may do computations in a coarse-to-fine manner, where the result is firstly regressed to an embedding around that of the right answer, and then converges to the final output based on the fine-grained information introduced by subsequent MLPs. 

Consolidating these findings, we can assert with some confidence that LLMs initially leverage attention heads to focus on operands (\A\ and \texttt{\{B\}}) and the operator, relaying this information to downstream MLPs. Over time, the MLPs progressively bolster \C\ and diminish the effect of confused answers, carrying out the calculation to final results.

\subsection{Precise SFT on Calculation-related Components.}
\label{exp:finetune}

\textbf{Experimental details.}
We evaluate precise SFT on four mathematical datasets (GSM8K \citep{gsm8k}, AddSub \citep{AddSub}, SingleEq \citep{SingleEQ}, SVAMP \citep{SVAMP}), and another two datasets (MMLU \citep{MMLU} and CSQA \citep{CSQA}) to evaluate the generic ability.
During training, we optimize the key components only and leave the other components unchanged.
We gather all training data from four mathematical datasets, and perform SFT updating on top $32$ key heads. Following \citep{Yu2023LanguageMA}, the gradient is rescaled by $\frac{H}{h}$, where $H$ is the number of all heads in each layer, $h$ is the number of updated heads in each layer. In practice, we train LLaMA2-7B and LLaMA2-13B with a learning rate of $2 \times 10^{-5}$ and a batch size of $128$ for $2$ epochs. The warm up ratio and weight decay are set as $0.02$ and $0.1$ by default, respectively. All experiments are conducted on $8$ NVIDIA A100 $80$GB GPUs.

\begin{table}[t!]
    \caption{Ablative experiments on the number of tunable components. The default setting is shown in \colorbox{midgrey}{gray}.
    }
    \label{tab:ablation-experiments}
    \resizebox{\columnwidth}{!}{
    \centering
    \begin{tabular}{@{}ccccc@{}}
    \toprule
    \multirow{2}{*}{\textbf{\makecell[c]{ Precise SFT \\ Setting}}} & \multicolumn{4}{c}{\textbf{\makecell[c]{Evaluation Metric}}} \\ \cmidrule(l){2-5} 
    & \makecell[c]{Train \\ Speed} & \makecell[c]{Tuned \\ Params.} & GSM8K & MMLU \\ \midrule
    top-8 heads   & 58sam./sec. & 0.017B & 25.4 & 45.1 \\
    top-16 heads        & 52sam./sec. & 0.033B & 26.5 & 45.8 \\
    \cellcolor{midgrey}top-32 heads        & \cellcolor{midgrey}50sam./sec. & \cellcolor{midgrey}0.067B & \cellcolor{midgrey}27.4 & \cellcolor{midgrey}46.4 \\
    top-48 heads        & 46sam./sec. & 0.101B & 27.4 & 46.4 \\
    top-64 heads        & 40sam./sec. & 0.134B & 27.3 & 45.5 \\ \midrule
    \makecell[c]{top-32 heads \\ + top-3 MLPs}  & 31sam./sec. & 0.473B & 28.0 & 45.2 \\ \bottomrule
    \end{tabular}
    }
\end{table}

\textbf{Precise SFT improves mathematical ability.}
Supervised Fine-Tuning (SFT) is an effective approach for augmenting the mathematical capabilities of models by fine-tuning all parameters within LLMs. We term this all-parameter fine-tuning as Full SFT for clarity, and adopt the same training settings as Precise SFT. Table~\ref{tab:exp:overall} presents the results of Full SFT and Precise SFT on the LLaMA2-7B and LLaMA2-13B models. Precise SFT effectively bolsters their mathematical capabilities, yielding an averaged increase of $15\%$ on four distinct mathematical datasets. It matches or even surpasses the improvements made by Full SFT. For example, Precise SFT outperforms Full SFT by $5.5\%$ on the SVAMP dataset and $2.8\%$ on GSM8K, underlining its superior ability to enhance the mathematical prowess of LLMs. Full SFT suffers from the trade-off between mathematical and general capabilities (about $5\%$ drops on MMLU and CSQA), while Precise SFT effectively maintains the model's original performance. A further advantage of Precise SFT is the drastic reduction in training time, attributed to the substantially fewer parameter adjustments required (less than $1\%$). It results in a time reduction of at least threefold on LLaMA2-7B and LLaMA2-13B. Overall, Precise SFT offers an effective direction for boosting mathematical abilities for LLMs.

\textbf{Ablative studies.}
The key issue with Precise SFT lies in determining the quantity and specific set of components to adjust. To demonstrate this, we experimented with varying numbers of heads and MLPs, with the results laid out in Table~\ref{tab:ablation-experiments}. We discovered that fine-tuning $32$ heads yields the best average improvement across different numbers of involved heads. We also compared experiments with the introduction of MLPs. We observed that as more MLPs are added, the mathematical capability improves by $2.1\%$, but the general performance will decrease by $1.5\%$ (results in Appendix \ref{app:MLP_ablation}). Overall, the top-$3$ MLPs yielded the best comprehensive results. However, even the introduction of a single MLP can reduce computational efficiency by $15\%$. How to more precisely fine-tune MLPs will be explored in our future work.

\textbf{More discussions.} 
The above results underscore the potential of employing interpretability tools to analyze the inner mechanism of LLMs and to enhance their specific capabilities. However, there are several areas that require deeper investigation: (i) Our primary experiments and discussions center around the LLaMA2 series. The results presented in Appendix \ref{app:PathP_LLM} demonstrate the potential for generalization across different LLMs, such as Mistral-7B \cite{jiang2023mistral}. For more rigorous considerations, it's necessary to perform specific adaptations on a broader range of LLMs.
(ii) This work mainly focuses on interpreting the fundamental ability of ``arithmetic calculation'', since it's universally shared across various levels of complexity for mathematical problems. The results in Appendix \ref{app:gsm8k} reveal that solving the math word problems requires a synergy of multiple skills including ``text comprehension'' and ``arithmetic calculation'', which is aligned with the findings in recent research \cite{human-cog}. It's imperative for continued research to investigate more complex mathematical problems. (iii) The potential of generalizing to more complex mathematical tasks like exponentiation (\eg, ``\A\ to the power of \B\ equals \_") has been validated in Appendix \ref{app:four_tasks_com}. An intriguing research direction would be to investigate the shared and distinct mechanisms across various mathematical tasks. 

%% file: tex/7_conclusion.tex
In this study, we have identified, analyzed, and fine-tuned the internal components responsible for the mathematical calculation capability of LLMs. The language models frequently involve sparse heads to particularly attend to operands and operators, and subsequent MLPs to work out answers. We apply the precise tuning on the calculation-related heads/MLPs for better
mathematical capabilities, with less impact on non-mathematical
tasks compared with tuning all parameters.
These findings contribute to a better understanding of the inner mechanism of LLMs.

%% file: tex/8_appendix.tex
\section{Appendix}

\appendix

\section{Templates}
\label{app:templates}

\begin{figure}[H]
\centering
\begin{tabular}{ |l|l| } 
\hline
\textbf{Addition} & \textbf{Subtraction} 
\\
\hline
\hline
\A\ + \B\ = \C\ & \A\ - \B\ = \C\ 
\\
\hline
\A\ plus \B\ equals to \C\ & \A\ minus \B\ equals to \C\ 
\\
\hline
The addition of \A\ and \B\ is \C\ & The difference of \A\ and \B\ is \C\ 
\\
\hline
The addition of \A\ and \B\ equals to \C\ & The difference of \A\ and \B\ equals to \C\ 
\\
\hline
The addition of \A\ and \B\ equals to \C\ & The difference of \A\ and \B\ equals to \C\ 
\\
\hline
Q: How much is \A\ plus \texttt{\{B\}}? A: & Q: How much is \A\ minus \texttt{\{B\}}? A:
\\
\hline
Q: What is \A\ plus \texttt{\{B\}}? A: & Q: What is \A\ minus \texttt{\{B\}}? A: 
\\
\hline
Q: What is the result of \A\ plus \texttt{\{B\}}? A: & Q: What is the result of \A\ minus \texttt{\{B\}}? A: 
\\
\hline
Q: What is the sum of \A\ and \texttt{\{B\}}? A: & Q: What is the difference of \A\ and \texttt{\{B\}}? A: 
\\
\hline
\hline
\textbf{Multiplication} & \textbf{Division} 
\\
\hline
\hline
\A\ * \B\ = \C\ & \A\ / \B\ = \C\ 
\\
\hline
\A\ times \B\ equals to \C\ & \A\ over \B\ equals to \C\ 
\\
\hline
The product of \A\ and \B\ is \C\ & The ratio of \A\ and \B\ is \C\ 
\\
\hline
The product of \A\ and \B\ equals to \C\ & The ratio of \A\ and \B\ equals to \C\ 
\\
\hline
The product of \A\ and \B\ equals to \C\ & The ratio of \A\ and \B\ equals to \C\ 
\\
\hline
Q: How much is \A\ times \texttt{\{B\}}? A: & Q: How much is \A\ over \texttt{\{B\}}? A:
\\
\hline
Q: What is \A\ times \texttt{\{B\}}? A: & Q: What is \A\ over \texttt{\{B\}}? A: 
\\
\hline
Q: What is the result of \A\ times \texttt{\{B\}}? A: & Q: What is the result of \A\ over \texttt{\{B\}}? A: 
\\
\hline
Q: What is the product of \A\ and \texttt{\{B\}}? A: & Q: What is the ratio of \A\ and \texttt{\{B\}}? A: 
\\
\hline
\end{tabular}
\caption{Templates used in this work follow the formations of ``Equation'', ``Statement'', ``Question-Answer''.}
\label{fig:base_templates}
\end{figure}

\begin{figure*}[htbp]
  \centering
    \includegraphics[width=0.85\linewidth]{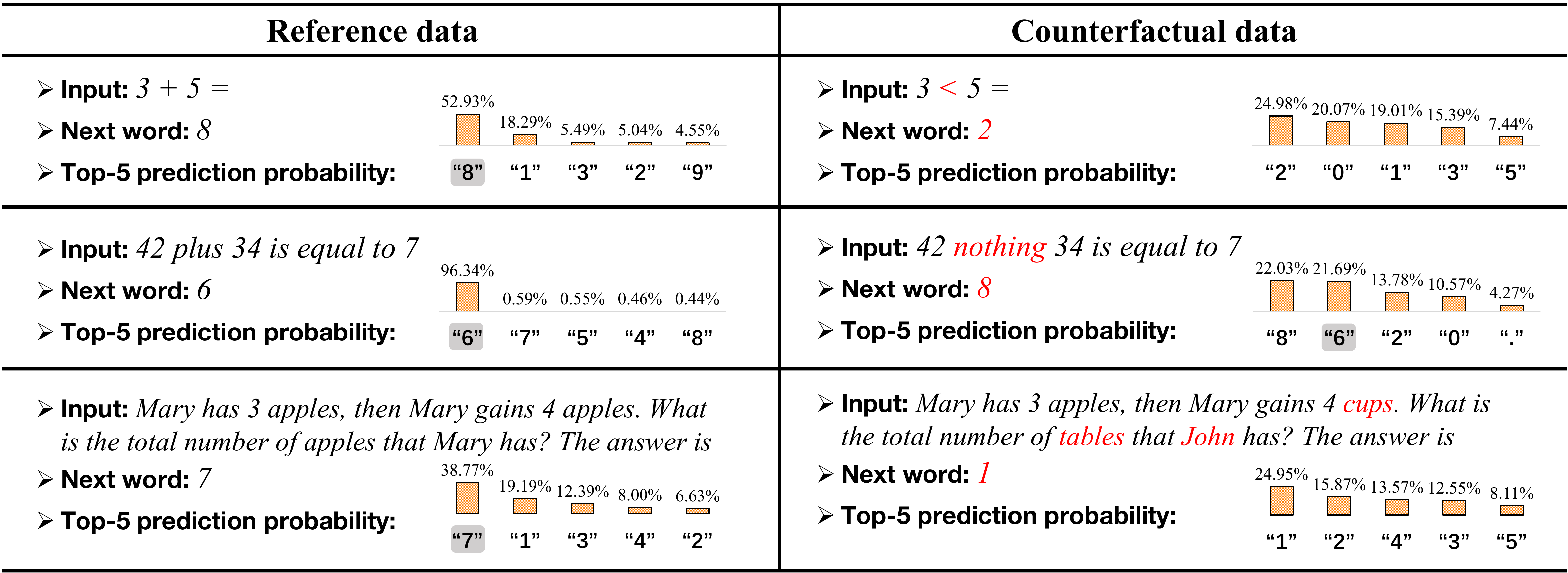}
    \caption{Examples of reference data (with addition logic) and counterfactual data (without addition logic). Given the input sentence, the results of next word prediction are provided by LLaMA2-7B.}
  \label{fig:xr_xc}
\end{figure*}

We have included a list of $36$ templates used in this work as shown in Figure~\ref{fig:base_templates}. 
All these templates share the same calculation logic. 
we sample the \texttt{<A>} and \texttt{<B>} from $\left \{ 1, \cdots, 9 \right \}$, since LLaMA2 tokenizes each digit individually (e.g., `$42$' is tokenized to `$4$' and `$2$'). Based on the above templates, we generate the sentences that the LLMs can predict the addition result \C\ correctly as the reference data $\xr$. We generate the counterfactual data $\xc$ following the principles depicted in Section~\ref{sec:related}, where we replace the words (\eg, ``plus'', ``minus'', ``times'', ``ratio'') with a randomly-selected term from the set $\left \{ \text{``none''}, \text{``nothing''}, \cdots, \text{``null''} \right \}$, and replace the operations (\eg, ``+'', ``-'', ``*'', ``/'') with a randomly-selected term from the set $\left \{ \text{``$<$''}, \text{``$>$''}, \cdots, \text{``@''} \right \}$. We show three cases in Figure \ref{fig:xr_xc} with the inspection into the top-5 prediction probability of LLaMA2-7B.
Moreover, in Figure \ref{fig:table_templates}, we also construct several different types of linguistic meanings for the addition task: ``time span'' and ``object accumulation''. 
For the templates 1-8 of ``time span'', we sample from a curated list of common words. For example, we select \texttt{<EVENT>} from $\left \{ \text{``war''}, \text{``conflict''}, \cdots, \text{``project''} \right \}$\footnote{We empirically find that the specific choice of words does not affect the results, as long as they meet similar semantics.}, \texttt{<VERB>} from $\left \{ \text{``last''}, \text{``span''}, \cdots, \text{``extend''} \right \}$, \texttt{<MONTH>} from $\left \{ \text{``Jan.''}, \text{``Feb.''}, \cdots, \text{``Dec.''} \right \}$, and \texttt{<YYY>} from $\left \{ 100, \cdots, 199 \right \}$.
For the templates 9-12 of ``object accumulation'', we sample \texttt{<OBJECT>} from $\left \{ \text{``apple''}, \text{``orange''}, \cdots, \text{``pear''} \right \}$, \texttt{<VERB>} from $\left \{ \text{``get''}, \text{``obtain''}, \cdots, \text{``acquire''} \right \}$, and each \texttt{<NAME>} was randomly selected from a pool of $100$ English first names.

\begin{figure}[H]
\centering
\begin{tabular}{ |l| } 
\hline
1. The \texttt{<EVENT>} \texttt{<VERB>} \A\ years from the year \texttt{<YYY>}\B\ to the year \texttt{<YYY>}\C \\
\hline 
2. The \texttt{<EVENT>} \texttt{<VERB>} \A\ years from \texttt{<YYY>}\B\ to \texttt{<YYY>}\C \\
\hline
3. The \texttt{<EVENT>} \texttt{<VERB>} \A\ days from \texttt{<MONTH>} \B\ to \texttt{<MONTH>} \C  \\
\hline
4. The \texttt{<EVENT>} will \texttt{<VERB>} \A\ days from \texttt{<MONTH>} \B\ to \texttt{<MONTH>} \C \\
\hline
5. The \texttt{<EVENT>} \texttt{<VERB>} \A\ hours from \B\ pm to \C
\\
\hline
6. The \texttt{<EVENT>} will \texttt{<VERB>} \A\ hours from \B\ pm to \C \\
\hline
7. The \texttt{<EVENT>} \texttt{<VERB>} \A\ hours from \B\ am to \C 
\\
\hline
8. The \texttt{<EVENT>} will \texttt{<VERB>} \A\ hours from \B\ am to \C 
\\
\hline
9. \texttt{<NAME>} has \A\ \texttt{<OBJECT>}, then \texttt{<NAME>} \texttt{<VERB>} \B\ \texttt{<OBJECT>}. \\ What's the total number of \texttt{<OBJECT>} that \texttt{<NAME>} has? The answer is \C
\\
\hline
10. \texttt{<NAME>} \texttt{<VERB>} \A\ \texttt{<OBJECT>}, and \texttt{<NAME2>} \texttt{<VERB>} \B\ \texttt{<OBJECT>}. \\ What's the total number of \texttt{<OBJECT>} that they \texttt{<VERB>}? The answer is \C
\\
\hline
11. \texttt{<NAME>} has \A\ \texttt{<OBJECT>}, and \texttt{<NAME2>} has \B\ \texttt{<OBJECT>}. \\ What's the total number of \texttt{<OBJECT>} that they have? The answer is \C
\\
\hline
12. \texttt{<NAME>} \texttt{<VERB>} \A\ \texttt{<OBJECT>} yesterday, and \texttt{<NAME>} \texttt{<VERB>} \B\ \texttt{<OBJECT>} today. \\ What's the total number of \texttt{<OBJECT>} that \texttt{<NAME>} \texttt{<VERB>}? The answer is \C
\\
\hline
\end{tabular}
\caption{Additional templates used in the addition task, involve different linguistic meanings like ``time span'' (1-8) and ``object accumulation'' (9-12).}
\label{fig:table_templates}
\end{figure}

\section{Evaluate the Effect of Attention Heads.}
\label{app:general_pp}
\begin{figure}[htbp]
  \centering
    \includegraphics[width=1.0\linewidth]{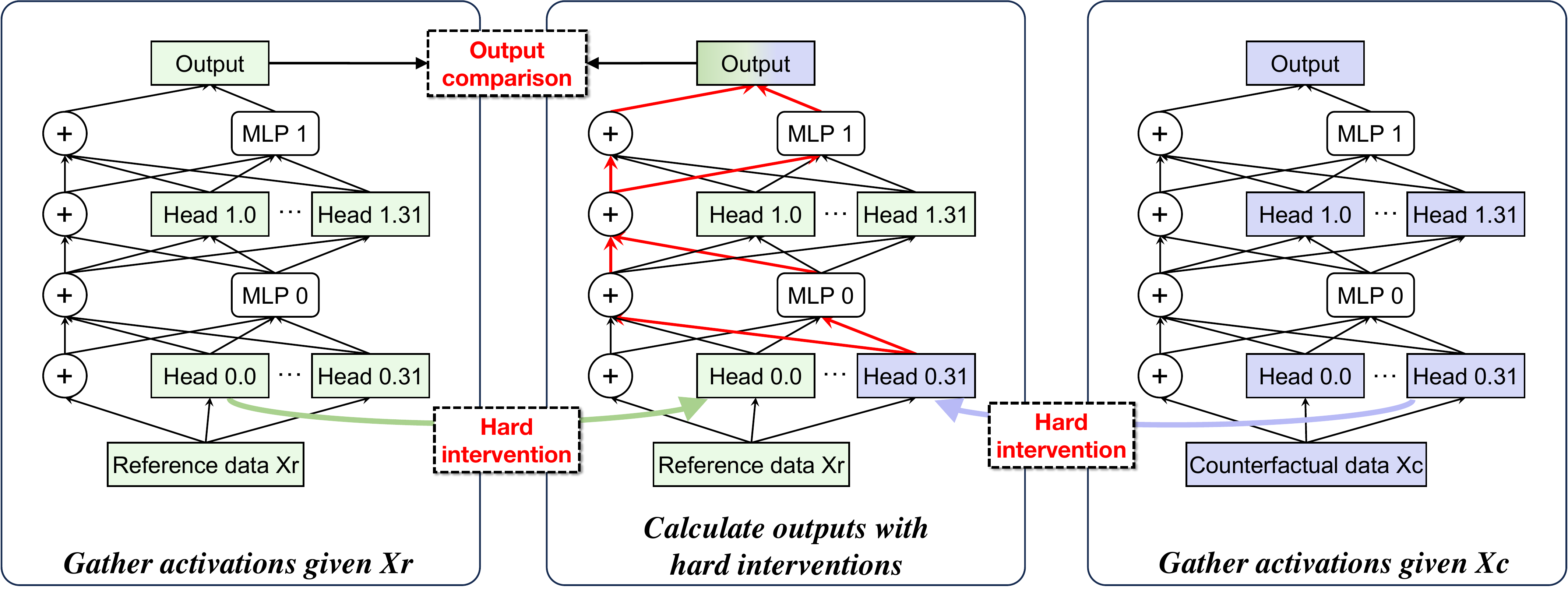}
    \caption{A case illustration of the method ``path patching''. It measures the importance of forward paths (\ie, the red lines that originate from Head $0.31$ to Output) for the two-layer transformer in completing the task on reference data. 
    }
  \label{fig:method_app}
\end{figure}

\textbf{Path Patching.}
To discover the cause of the predicted answer, we employ the causal intervention technique known as \textit{path patching} \citep{localizing, inter-IOI}. This approach is highly effective in analyzing the causal relationship between two computation nodes (Sender $\xrightarrow{}$ Receiver).
This helps us determine whether Sender is the cause of Receiver, and the connections between them are important for the model in implementing the task. 

Specifically, the entire process of path patching is shown in Figure~\ref{fig:method_app}, where the node pair Sender $\xrightarrow{}$ Receiver is set as Head $0.31$ $\xrightarrow{}$ Output.
Firstly, given reference data $\xr$ and counterfactual data $\xc$, the activations of all heads are gathered for preparation of the later perturbation.
Then, we do a hard intervention on the Head $0.31$ that is perturbated to its activation on $\xc$, where the effect will be further propagated to the Ouput node along with a set of paths $\mathcal{P}$.
To ensure an independent observation of the impact from the Head $0.31$, $\mathcal{P}$ comprises the forward pathways through residual connections and MLPs except for the other attention heads (\eg, Head $0.0, \cdots, 0.30, 1.0, \cdots, 1.31$). 
Thus we do a hard intervention on the other heads by freezing their activations on $\xr$.
Finally, we obtain the final output logits to measure the impact of this perturbation.
If there is a significant change in final logits, then the patched paths: Sender $\xrightarrow{}$ Receiver are essential for the model in completing the task. 

In this work, to identify the important heads contributing to the calculation task, we scan through all heads  as the Sender node denoted by $h$, and set the Receiver node as output $logits$, and measure the changes in the output logit of ground-truth token \C. 
Pathways $h \to logits$ that are critical to the model's computation should induce a large drop in the logit of token \C\ after patching. Notably, since the residual operations and MLPs compute each token separately \citep{elhage2021mathematical}, patching the head output at the END position (\ie, the position of the last token in the input sentence) is enough to measure the effects on the next token prediction.

\section{More Results of Other LLMs.}
\label{app:PathP_LLM}

\begin{figure}[htbp]
  \centering
    \includegraphics[width=0.9\linewidth]{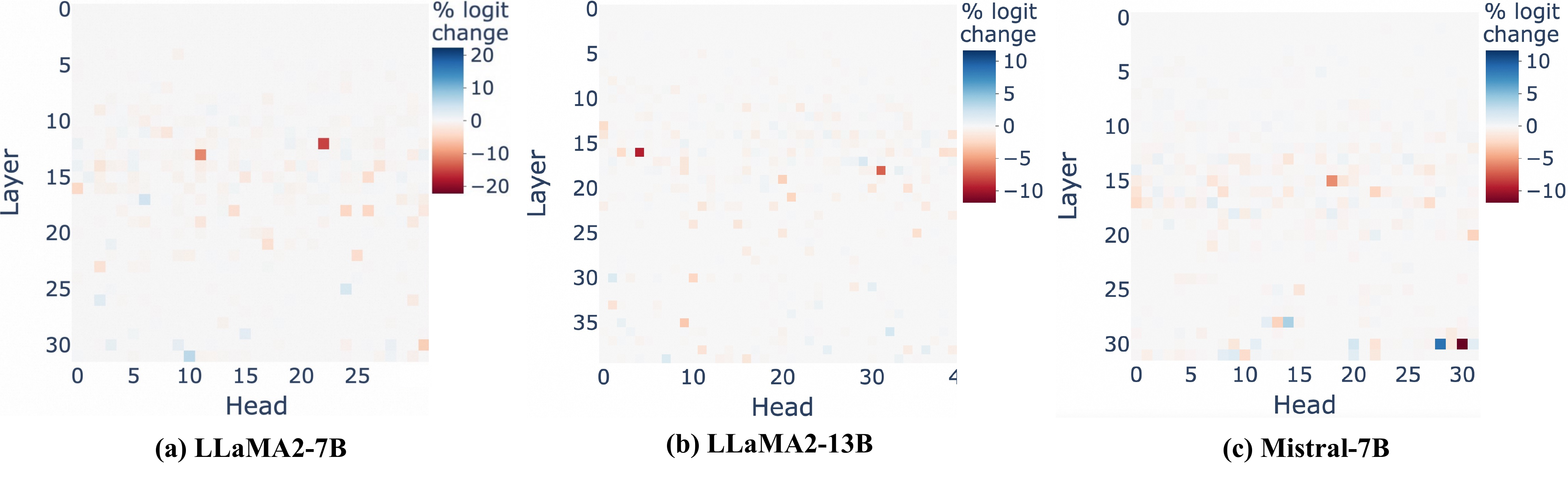}
    \vspace{-3mm}
    \caption{Comparison of the results of path patching experiments on LLaMA2-7B, LLaMA2-13B, and Mistral-7B \cite{jiang2023mistral} across four mathematical tasks. For each head/MLP, a darker color indicates a larger logit difference from the original model before patching.
    }
  \label{app:7b_13b_compare}
\end{figure}

\textbf{Key Component Identification.} In Figure \ref{app:7b_13b_compare}, we further report the results of key components identification of other models (\eg, LLaMA2-13B and Mistral-7B). For example, LLaMA2-13B comprises $40$ layers and $40$ attention heads per attention layer. The three models of different size exhibit similar phenomena that the calculation-related key heads (\eg, $16.4$, $18.31$) are distributed sparsely in the middle layers.

\begin{figure}[htbp]
  \centering
    \includegraphics[width=\linewidth]{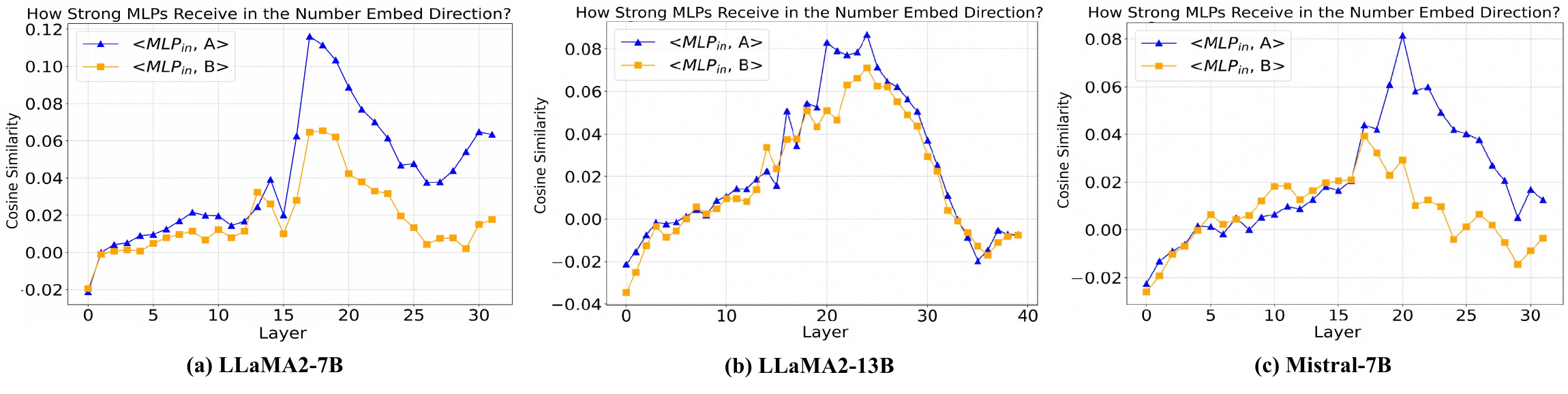}
    \vspace{-3mm}
    \caption{We investigate the projection of each MLP layer input ($MLP_{in}$) along the direction of number token \A, \B, respectively.
    }
  \label{app:other_model_receive}
\end{figure}

\begin{figure}[htbp]
  \centering
    \includegraphics[width=\linewidth]{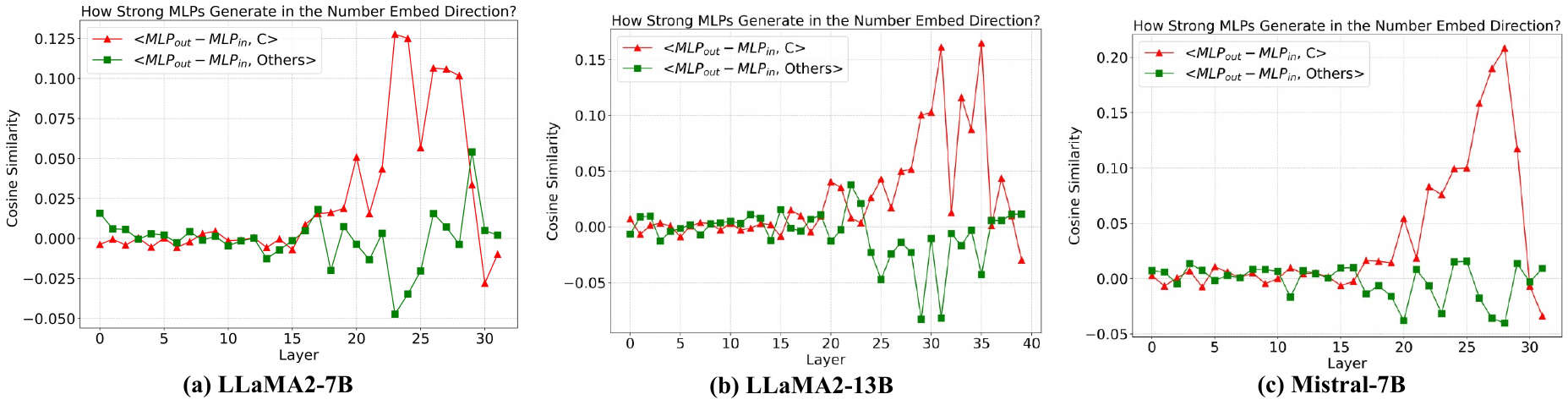}
    \vspace{-3mm}
    \caption{We investigate the projection of each MLP layer ($MLP_{out}$-$MLP_{in}$) along the direction of number token \C\ (\ie, right answer) and other tokens (\ie, wrong answer).
    }
  \label{app:other_model_generate}
\end{figure}

\textbf{Key MLPs Behavior.} 
In Figure \ref{app:other_model_receive}, the similarities of MLP input and number operands \A/\B\ across all models demonstrate ascending and descending trends. Specifically, the pivotal points for these trends, delineated as ($start$-$inflection$-$end$), are as follows: ($13$-$18$-$28$) for LLaMA2-7B, ($13$-$18$-$35$) for LLaMA2-13B, and ($13$-$20$-$28$) for Mistral-7B.
In Figure \ref{app:other_model_generate}, the similarities of $MLP_{out}$-$MLP_{in}$ and right answer \C\ show a pattern of initial stabilization followed by an increase. The critical points for LLaMA2-7B/LLaMA2-13B/Mistral-7B are again ($13$-$18$-$28$), ($13$-$18$-$35$), and ($13$-$20$-$28$). The inflection points in both Figure \ref{app:other_model_receive} and Figure \ref{app:other_model_generate} are nearly identical, indicating consistent trend shifts across the models. It helps to verify that LLMs initially leverage attention heads then relaying information to downstream MLPs, to progressively carry out the calculation to final results.
Furthermore, the above findings appear to be general and robust across different LLMs, not limited to a specific model.

\section{Key Component Location across Calculation Tasks.}
\label{app:four_tasks_com}
\begin{figure*}[htbp]
  \centering
    \includegraphics[width=0.95\linewidth]{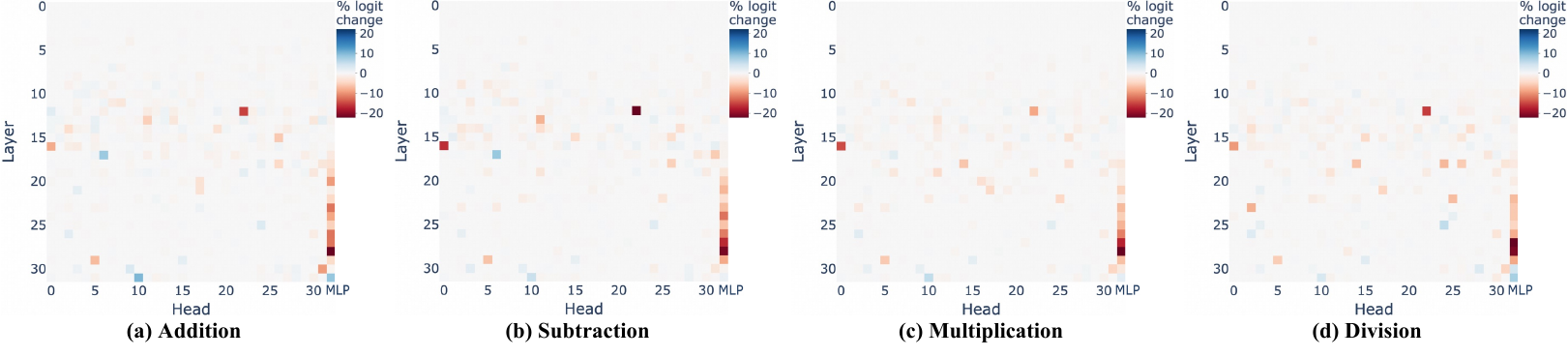}
    \caption{We conduct path patching experiments on LLaMA2-7B across four mathematical tasks, by searching for each head and MLP directly affecting the logit of the right answer. The last column denotes the path patching results of MLPs. For each head/MLP, a darker color indicates a larger logit difference from the original model before patching. }
  \label{app:four_tasks}
\end{figure*}

We investigate the location of key components for each calculation task individually, as shown in Figure \ref{app:four_tasks}. 
The discovered key heads could be shared across four tasks, which are sparsely distributed in the middle layers. Specifically, when examining subtraction and addition tasks, we could summarize two insightful symmetries between them. The identified key heads of two tasks are almost the same, albeit with different magnitude of the effect. This phenomenon could reveal the symmetry of key head ``\textit{location}'' in addition and subtraction. 
Moreover, the tasks of multiplication and division exhibit a greater number of key heads compared to the tasks of addition and subtraction. We assume 
it could be attributed to their more intricate operations within multiplication and division.

\textbf{Generalize to other calculation operations.} We conduct the experiments of key head identification and validation following Section \ref{exp:discover}. We generate the samples including the exponentiation operation as the reference data $\xr$. Then we generate the counterfactual data $\xc$ following the principles introduced in Section \ref{sec:identify} to exclude the exponentiation logic. 

\begin{table}[h]
\centering
\caption{Key head identification on the exponentiation task.}
\begin{tabular}{l|c|c}
\hline
Templates & Key Heads [\eg, (Layer, Head)] & Knockout Accuracy \\
\hline
\makecell[l]{$\xr$: ``\A\ $\mathbf{\wedge}$ \B\ = \_ '' \\ $\xc$: ``\A\ $\mathbf{<}$ \B\ = \_ ''} & [(11, 8), (12, 22), (13, 11), (14, 2), (15, 15)]  & $-66\%$  \\
\hline
\makecell[l]{$\xr$: ``\A\ to the \textbf{power} of \B\ equals \_ '' \\ $\xc$: ``\A\ to the \textbf{none} of \B\ equals \_ ''} & [(11, 8), (12, 22), (13, 11), (14, 2), (15, 15)]  & $-62\%$  \\
\hline
\end{tabular}
\end{table}

The results reveal the potential of generalizing to more complex mathematical operations: (i) Five key heads are identified based on the newly generated $\xr$ and $\xc$. We find that the heads ($11$, $8$) and ($14$, $2$) mainly attend to the operators ``$\wedge$'', ``power'', while the heads ($12$, $22$), ($13$, $11$), ($15$, $15$) mainly attend to the input operands \A\ and \B. (ii) Knocking out the key heads, identified by both templates, leads to significantly impacts (over $60\%$) on model performance.

\section{Generalize to More Complex Scenarios.}
\label{app:gsm8k}
We conduct experiments on the more complex scenario using the dataset GSM8K \cite{gsm8k}. At first, we create new reference data $\xr$ and counterfactual data $\xc$. Following the idea of methodology proposed in Section \ref{sec:identify}, we convert the question in GSM8K to obfuscate the semantic elements that necessitate calculation, while ensuring that the alterations to the text are minimal. An example is shown below:

\begin{itemize}
    \item GSM8K $\xr$: ``On a $16$ GB (gigabyte) capacity USB drive, $50\%$ is already busy. \textbf{Calculate the number} of gigabytes still available.''  
    \item GSM8K $\xc$: ``On a $16$ GB (gigabyte) capacity USB drive, $50\%$ is already busy. \textbf{Describe the location} of gigabytes still available.''  
\end{itemize}

Then, we conduct the experiments of key head identification and validation following the experimental setting in Section \ref{exp:discover}. As a result, $60\%$ of the key heads are overlapped with the key heads identified based on our original data. Moreover, knocking out the newly-identified key heads leads to a $65\%$ accuracy drop on GSM8K, confirming their importance even in complex scenarios. 

\begin{table}[h]
\centering
\caption{Comparison of the key heads identified on our generated data in Figure \ref{fig:base_templates} and the dataset GSM8K \cite{gsm8k}.}
\begin{tabular}{l|c|c}
\hline
Dataset & Top-$10$ Key Heads [\eg, (Layer, Head)] & \makecell[c]{Knockout \\Accuracy} \\
\hline
Ours & [(12, 22), (13, 11), (16, 0), (15, 26), (18, 26), (18, 24), (30,31), (14, 27), (22, 25), (11, 8)] & $-69\%$  \\
\hline
GSM8K & [(19, 6), (11, 8), (12, 22), (14, 31), (13, 11), (22, 25), (16, 0), (21, 17), (15, 26), (29, 5)] & $-65\%$  \\
\hline
\end{tabular}
\end{table}

Furthermore, only knocking out the $6$ overlapping heads brings in $-56\%$ and $-52\%$ on our generated data and GSM8K, respectively. It shows these heads are both important in two scenarios. If knocking out the $4$ non-overlapping heads identified by GSM8K only, it has a negligible effect on our generated data ($-2\%$) but apparently affects on GSM8K ($-26\%$). It reveals the significance of these $4$ heads specific to more complex reasoning mathematical problems. We further investigate the attention patterns of the $4$ non-overlapping heads, and find that these heads mainly attend to text tokens. For example, the head ($29$, $5$) attends to ``.'', and the head ($19$, $6$) attends to ``GB''. In contrast, the $6$ overlapping heads mainly attend to the number operands and operators. For example, the head ($13$, $11$) attends to input operands ``50'', and the head ($11$, $8$) attends to the operator ``\%''.

Recent research \cite{human-cog} has shown that solving the math word problems requires a synergy of multiple skills including `text comprehension' and `arithmetic calculation'. This is aligned with the phenomena of ``the $4$ non-overlapping heads attend to text tokens (\ie, `text comprehension'), while the $6$ overlapping heads attend to number operands and operators (\ie, `arithmetic calculation')''. In this work, we focus on the skill of arithmetic calculation as it's a fundamental ability universally shared across various levels of complexity for mathematical problems. It's imperative for continued research to develop a more holistic understanding of the intricate reasoning capacities.

To further investigate whether the model's deficiencies stem from a lack of mathematical abilities or a broader impairment in language processing, we evaluate LLaMA2-7B with key heads kept normal and knocked out on MMLU-Humanities benchmark \cite{MMLU}. The comparative performance was $42.9\%$ for models with the key heads intact versus $42.6\%$ for the knockout models. This negligible difference ($-0.3\%$) suggests that the knockout of these heads does not significantly impact general language abilities.

\section{More Attention Pattern Cases.}
\label{app:attn_pattern}
\begin{figure*}[htbp]
  \centering
\subfigure[Head $14.2$]{
    \includegraphics[width=0.23\linewidth]{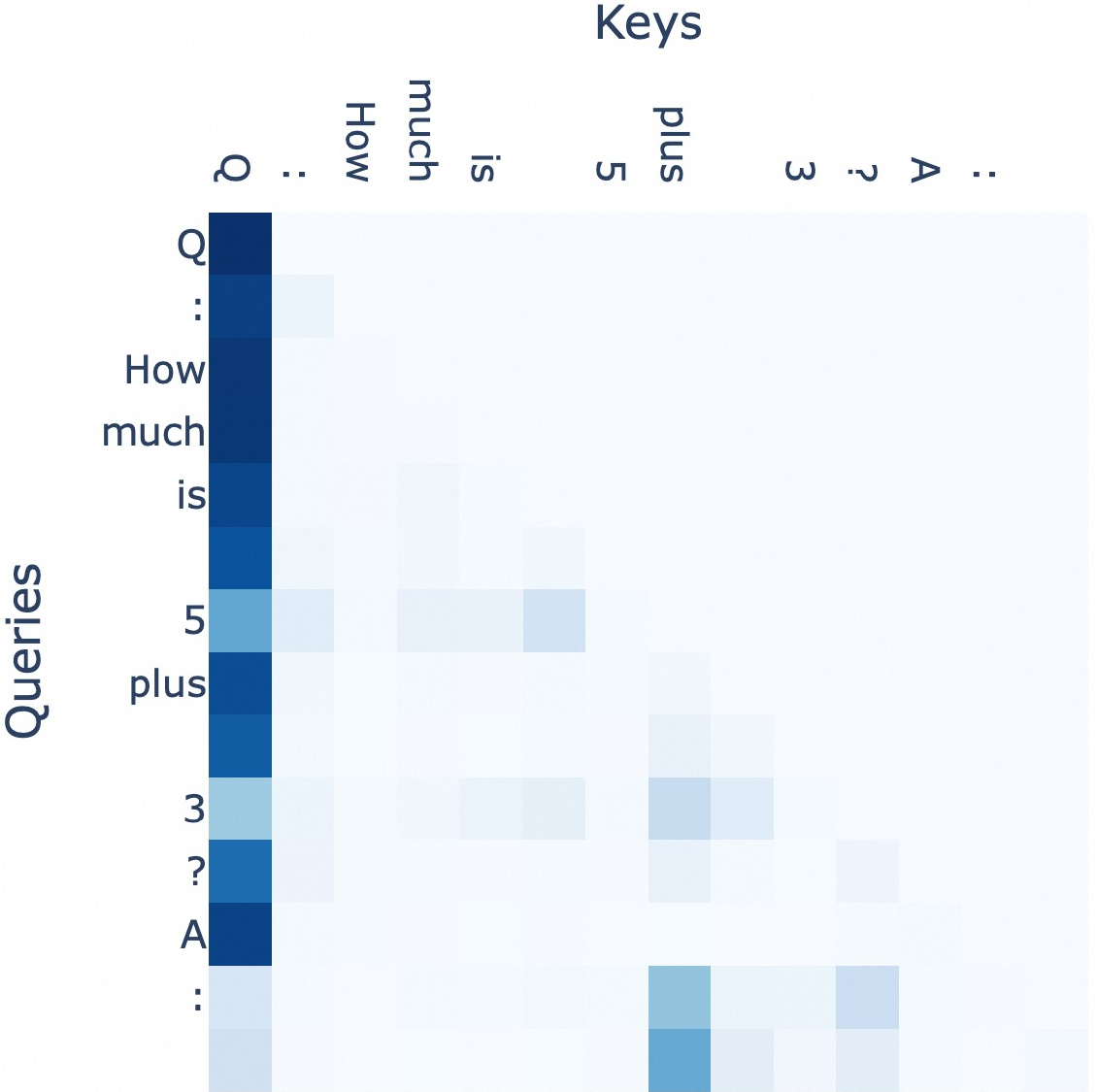}
  }
  \subfigure[Head $14.2$]{
    \includegraphics[width=0.23\linewidth]{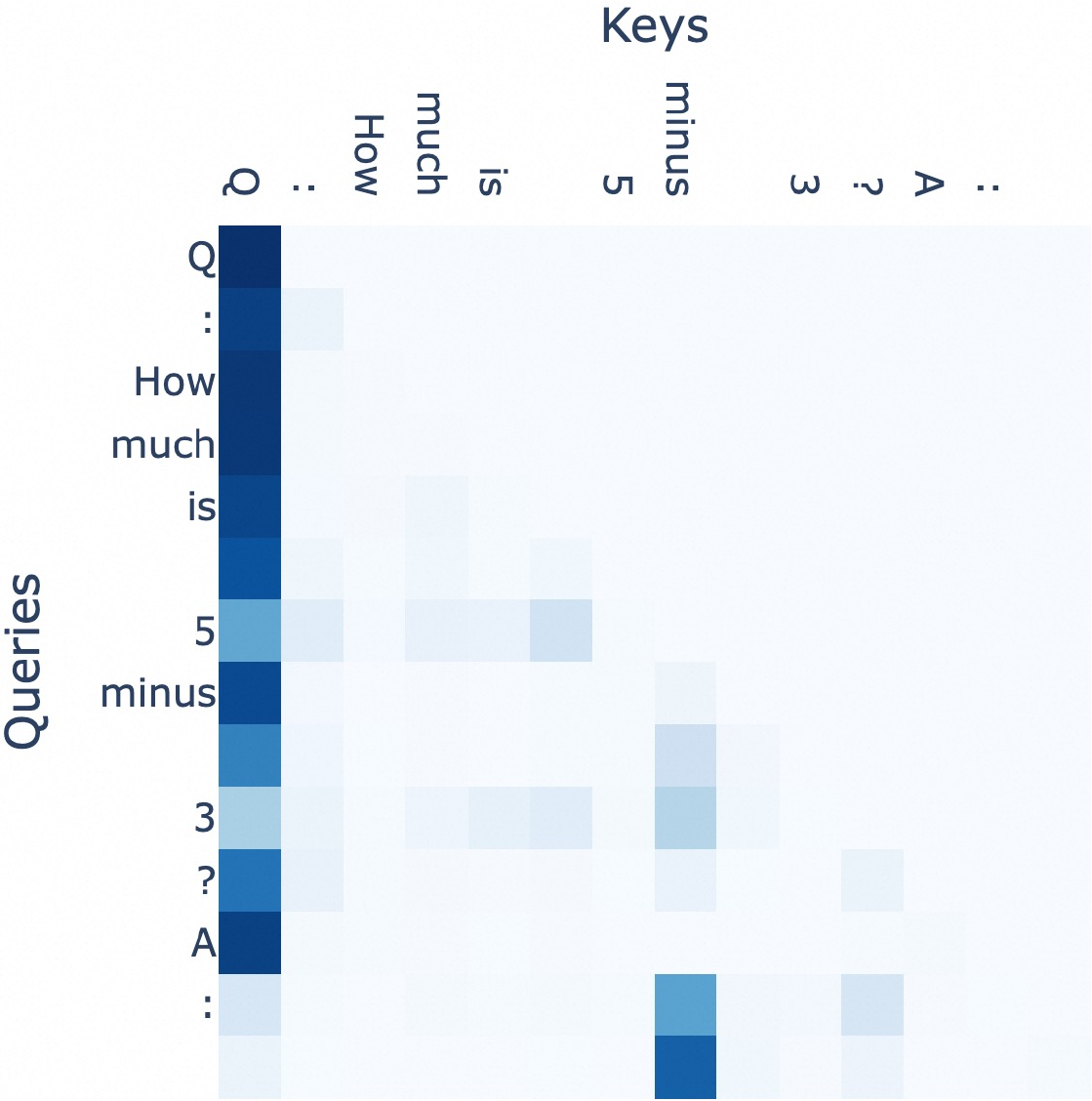}
  }
  \subfigure[Head $14.2$]{
    \includegraphics[width=0.23\linewidth]{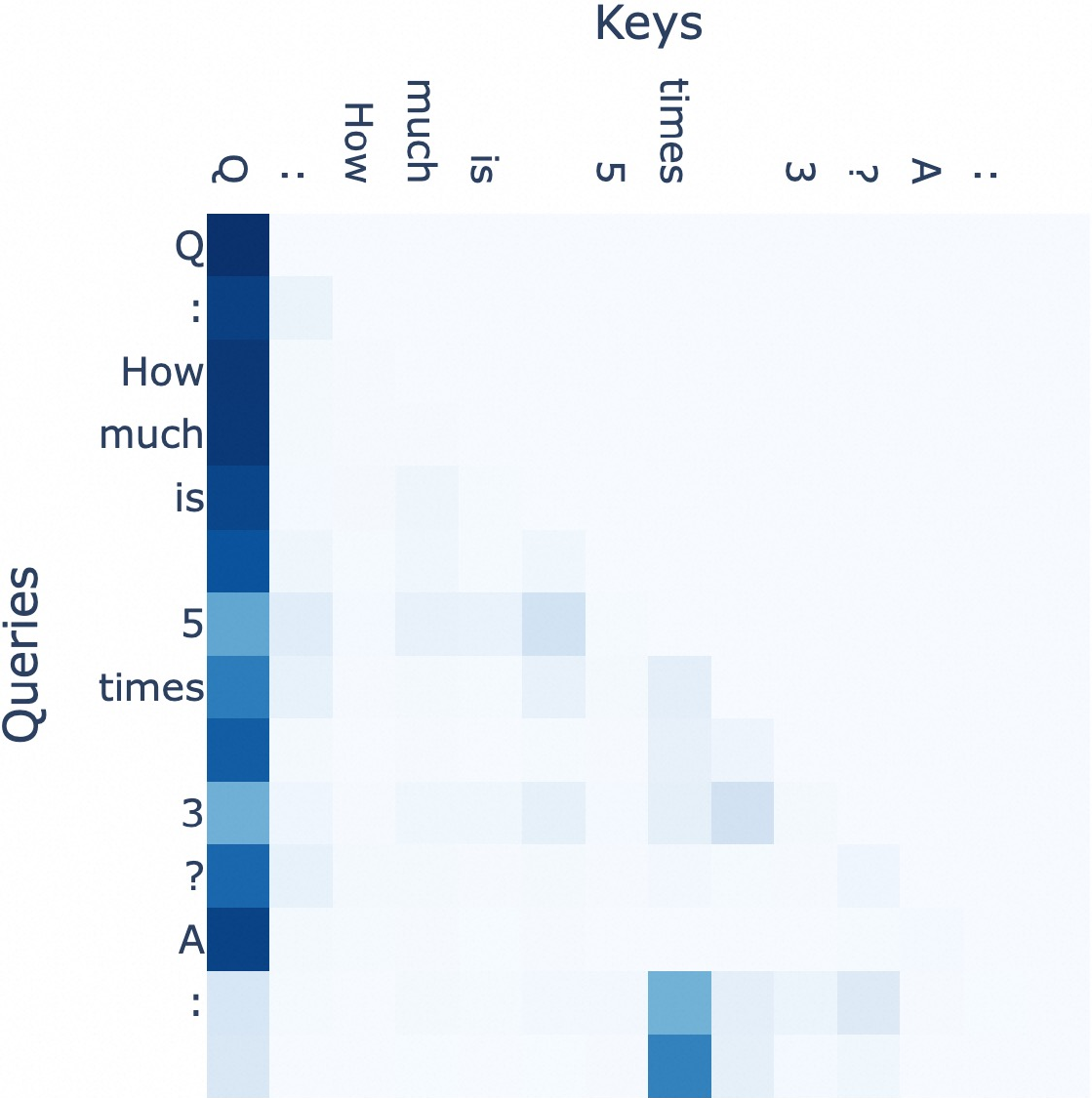}
  }
  \subfigure[Head $14.2$]{
    \includegraphics[width=0.23\linewidth]{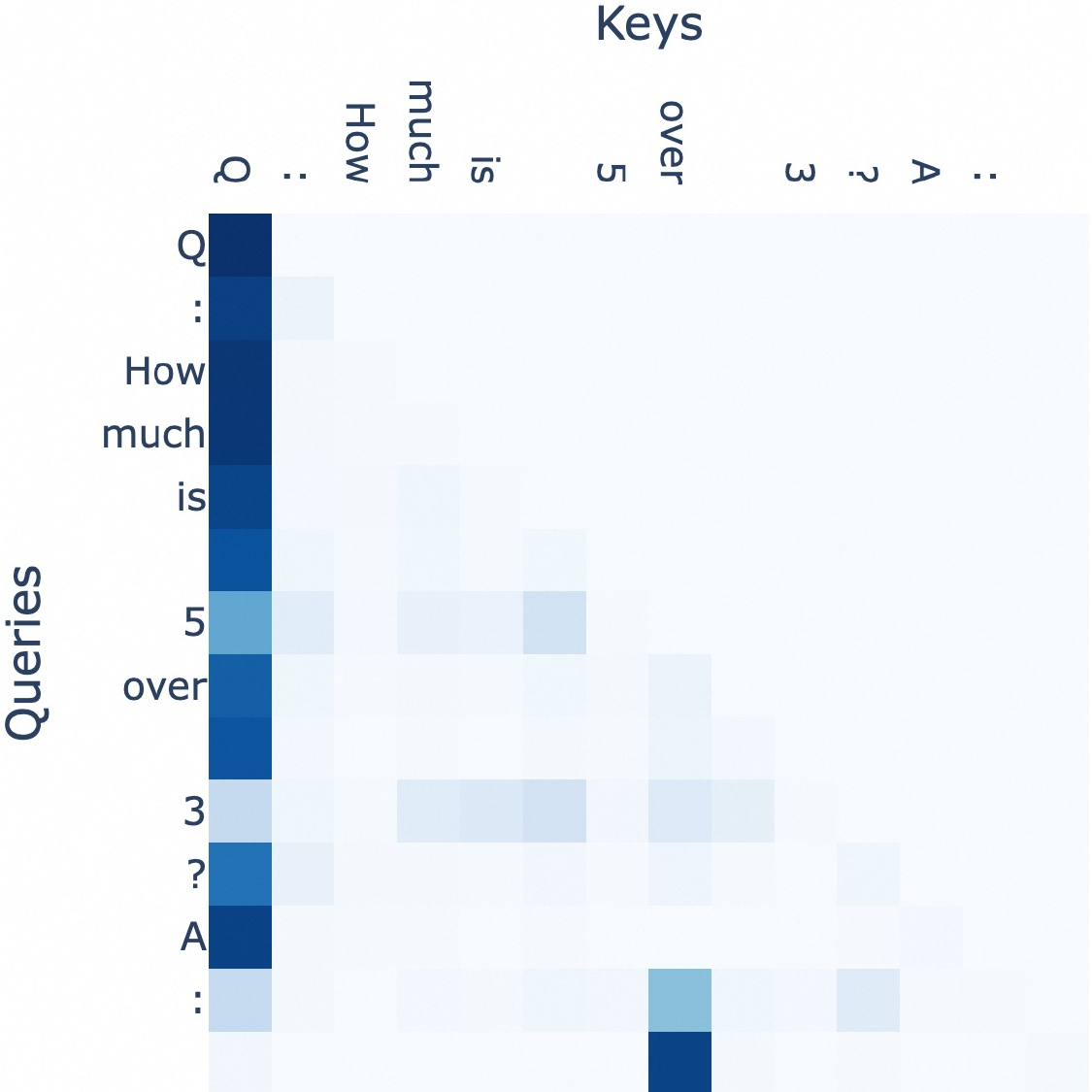}
  }
  \vspace{-3mm}
  \caption{The attention patterns of the key head $14.2$, which mainly attend to the operator-related tokens, \eg, ``plus'', ``minus'', ``times'', ``over''.}
   \label{app:attn_operator} 
\end{figure*}

We show the attention patterns of 
the operator-attended heads (\eg, $14.2$) in Figure \ref{app:attn_operator} that could attend to the tokens of ``plus'', ``minus'', ``times'', and ''over'', across different sentences.

\section{Ablation Study of Precise SFT on MLPs.}
\label{app:MLP_ablation}
We further investigate the influence of different number of tuned MLPs in Table \ref{app:MLP-experiments}. It reveals that the tuning more MLPs could lead to a performance decrease on MMLU and more training time, while imrove the performance on math dataset GSM8K. 

\begin{table*}[htbp]
    \caption{Ablative experiments on the number of tunable MLPs.
    }
    \label{app:MLP-experiments}
    \centering
    \begin{tabular}{@{}ccccc@{}}
    \toprule
    \multirow{2}{*}{\textbf{Precise SFT Setting}} & \multicolumn{4}{c}{\textbf{\makecell[c]{Evaluation Metric}}} \\ \cmidrule(l){2-5} 
    & \makecell[c]{Train \\ Speed} & \makecell[c]{Tunable \\ Params.} & GSM8K & MMLU \\ \midrule
    top-32 heads & 50sam./sec. & 0.067B & 27.4 & 46.4 \\
    \midrule
    top-32 heads + top-1 MLP  & 44sam./sec. & 0.202B & 27.5 & 46.0 \\
    \midrule
    top-32 heads + top-2 MLPs  & 38sam./sec. & 0.338B & 27.7 & 45.7 \\
    \midrule
    top-32 heads + top-3 MLPs  & 31sam./sec. & 0.473B & 28.0 & 45.2 \\
    \midrule
    top-32 heads  + top-6 MLPs  & 26sam./sec. & 0.879B & 28.2 & 44.9 \\
    \midrule
    top-32 heads + all MLPs  & 19sam./sec. & 4.396B & 29.2 & 43.9 \\ \bottomrule
    \end{tabular}
\end{table*}

\section{Calculation in Computer vs LLMs.}
\label{app:computer_vs_llms}
\begin{figure}[htbp]
  \centering
    \includegraphics[width=0.5\linewidth]{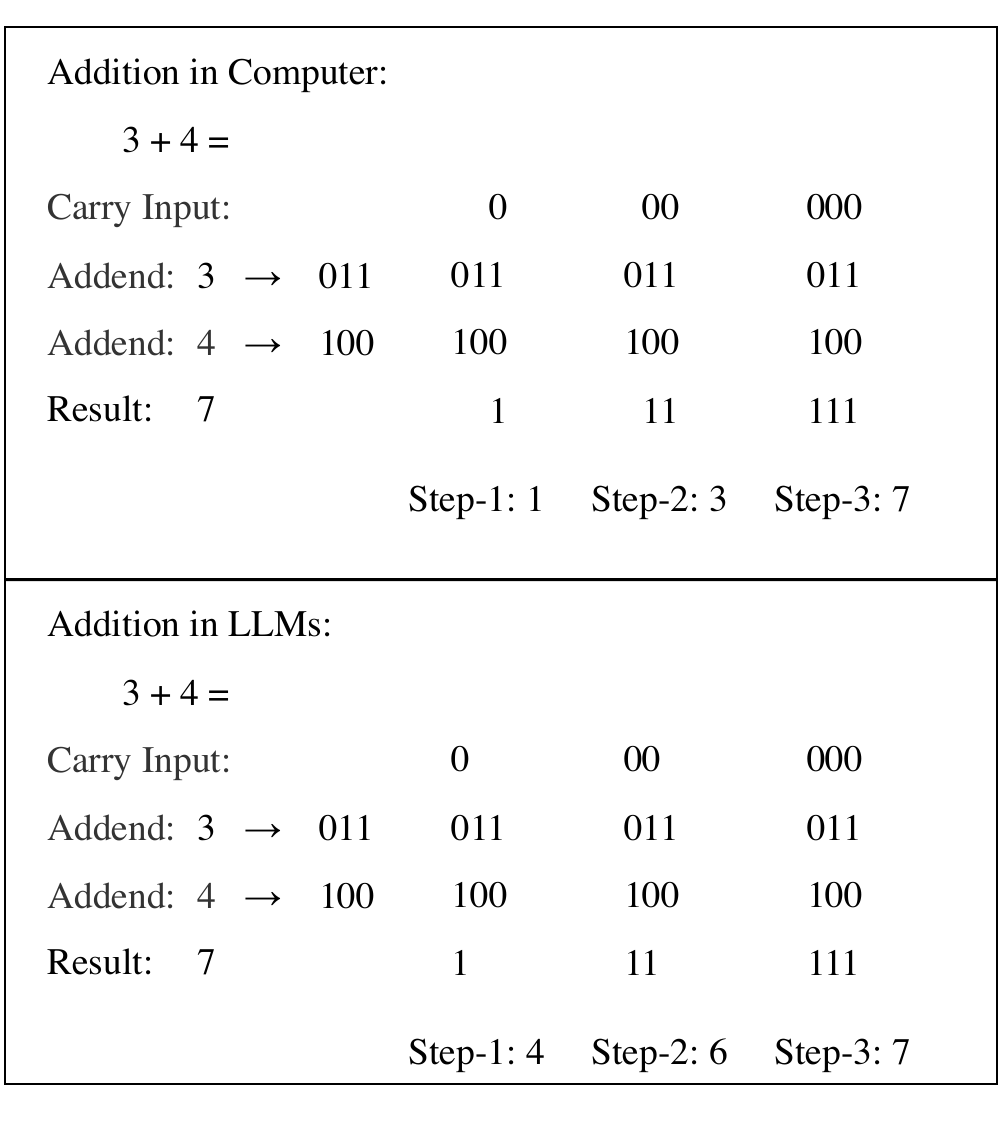}
    \caption{The addition calculation process in computer and in LLMs.
    }
  \label{fig:computer_vs_llms}
\end{figure}